\documentclass{article}

    \PassOptionsToPackage{numbers, compress}{natbib}

    \usepackage[preprint]{neurips_data_2022}

\usepackage[utf8]{inputenc} 
\usepackage[T1]{fontenc}    
\usepackage{hyperref}       
\usepackage{url}            
\usepackage{booktabs}       
\usepackage{amsfonts}       
\usepackage{nicefrac}       
\usepackage{microtype}      
\usepackage{xcolor}         

\usepackage{multirow}
\usepackage{makecell}
\usepackage{graphicx}
\usepackage{enumitem}

\usepackage{arydshln}

\usepackage{framed}
\usepackage{enumitem}
\usepackage{adjustbox}

\newif\ifcomments
\commentstrue
\ifcomments
    \providecommand{\sameer}[2][]{{\protect\color{violet}{[Sameer:\textbf{#1} #2]}}}
    \providecommand{\daniel}[2][]{{\protect\color{red}{[Daniel:\textbf{#1} #2]}}}
    \providecommand{\zhouhang}[2][]{{\protect\color{orange}{[Zhouhang:\textbf{#1} #2]}}}
    \providecommand{\jonathan}[2][]{{\protect\color{blue}{[Jonathan:\textbf{#1} #2]}}}
    \providecommand{\kalyani}[2][]{{\protect\color{pink!50!black}{[Kalyani:\textbf{#1} #2]}}}
    \providecommand{\wencong}[2][]{{\protect\color{gray}{[Wencong:\textbf{#1} #2]}}}
\else
    \providecommand{\sameer}[2][]{}
    \providecommand{\daniel}[2][]{}
    \providecommand{\zhouhang}[2][]{}
    \providecommand{\jonathan}[2][]{}
    \providecommand{\kalyani}[2][]{}
    \providecommand{\wencong}[2][]{}
\fi

\newcommand{\eat}[1]{}

\newcommand{\trim}[1]{}

\title{TCAB: A Large-Scale Text Classification Attack Benchmark\eat{\sameer{remove `Dataset'?}}}

\newcommand{\ucimark}[0]{\footnotemark[2]}
\newcommand{\uomark}[0]{\footnotemark[3]}
\newcommand{\ucsdmark}[0]{\footnotemark[4]}
\newcommand{\auth}[1]{\textbf{#1}}
\newcommand{\authspace}[0]{\hspace{7mm}}
\author{%
  Kalyani Asthana\ucimark$^{\hspace{1mm}}$\thanks{Corresponding author.}\authspace Zhouhang Xie\ucsdmark \authspace Wencong You\uomark \authspace Adam Noack\uomark\\
  \And
  \auth{Jonathan Brophy\uomark \authspace Sameer Singh\ucimark \authspace Daniel Lowd\uomark}\\
    \\
  \ucimark\hspace{1mm} University of California Irvine \authspace \ucsdmark\hspace{1mm} University of California San Diego \authspace \uomark\hspace{1mm} University of Oregon\\
  \texttt{\{kasthana,sameer\}@uci.edu} \authspace \texttt{zhx022@ucsd.edu}\\ \texttt{\{wyou,anoac2k,jbrophy,lowd\}@cs.uoregon.edu}
}
\begin{document}

\maketitle

\begin{abstract}

We introduce the Text Classification Attack Benchmark (TCAB), a dataset for analyzing, understanding, detecting, and labeling adversarial attacks against text classifiers. TCAB includes 1.5 million attack instances, generated by twelve adversarial attacks targeting three classifiers trained on six source datasets for sentiment analysis and abuse detection in English. Unlike standard text classification, text attacks must be understood in the context of the target classifier that is being attacked, and thus features of the target classifier are important as well. 

TCAB includes all attack instances that are successful in flipping the predicted label; a subset of the attacks are also labeled by human annotators to determine how frequently the primary semantics are preserved.
The process of generating attacks is automated, so that TCAB can easily be extended to incorporate new text attacks and better classifiers as they are developed.
In addition to the primary tasks of detecting and labeling attacks, TCAB can also be used for attack localization, attack target labeling, and attack characterization. TCAB code and dataset are available at \url{https://react-nlp.github.io/tcab/}.
\end{abstract}

\section{Introduction}

Text classifiers have been under attack ever since spammers started evading spam filters, nearly 20 years ago~\cite{hulten04trends}. In recent years, however, attacking classifiers has become much easier to carry out. Many general-purpose attacks have been developed and are now available in standard, plug-and-play frameworks, such as TextAttack~\cite{morris2020textattack} and OpenAttack~\cite{zeng2020openattack}. The wide use of standard architectures and shared pretrained representations have further increased the risk of attacks by decreasing the diversity of text classifiers.

Our focus is on evasion attacks~\cite{barreno2006can},
in which an attacker attempts to change a classifier's prediction by making minor, semantics-preserving perturbations to the original input. To accomplish this, different adversarial attack algorithms employ different types of perturbations, search methods, and constraints. 
See Table~\ref{tab:attack_samples_brief} for some brief examples.

A common defense strategy is to make classifiers more robust, using algorithms with heuristic or provable guarantees on their performance~\cite{madry18towards,cohen19certified,liu2020robust,tan2020mind,dong2021towards,wang2020infobert,Ivgi2021AchievingMR,yoo-qi-2021-towards-improving,Wang2021AdversarialTW,miyato2017adversarial,Zhu2020FreeLB,jiang2020smart,jones2020robust,huang2019ibp,jia2019certified,shi2020verification}. However, these defenses are often computationally expensive or result in reduced accuracy. Therefore, as a complement to making classifiers more robust, we introduce the task of \emph{attack identification} --- automatically determining the adversarial attacks (if any) used to generate a given piece of text. The idea behind attack identification is that many attackers will use whatever attacks are most convenient, such as public implementations of attack algorithms, instead of developing new ones or implementing ones on their own. Thus, we can identify specific attacks instead of detecting or preventing \emph{all} possible attacks.

\begin{table}[tb]
\center
\caption{Attack Samples on SST-2}
\label{tab:attack_samples_brief}
\begin{tabular}{llcc}
\toprule
\textbf{Attack} & \textbf{Text} & \textbf{Label} & \textbf{Confidence} \\
\midrule
    Original
    & the acting is amateurish
    & Negative
    & 63.7\%
\vspace{0.1cm} \\
\hdashline \\
\vspace{-0.6cm} \\
    Pruthi~\cite{pruthi2019combating}
    & the acting is \textcolor{red}{amateirish}
    & Positive
    & 82.1\% \\
    DeepWordBug~\cite{gao2018deepwordbug}
    & the acting is \textcolor{red}{aateurish}
    & Positive
    & 91.2\%\\
    IGA~\cite{wang2019igawang}
    & the acting is \textcolor{red}{enthusiastic}
    & Positive
    & 62.3\%\\
\bottomrule
\end{tabular}
\end{table}

The primary focus of attack identification is attack \textit{labeling} --- determining which specific attack was used (or none). However, other valuable challenges exist under the umbrella of attack identification, such as 
\textit{attack target labeling}~(determining which model is being attacked), \textit{attack localization}~(identifying which parts of the text have been manipulated), and more~(see~\S\ref{sec:potential_tasks} for detailed subtask descriptions). These tasks give us information about how the attacks are being conducted, which can be used to develop defense strategies for the overall system, such as uncovering malicious actors behind misinformation or abuse campaigns on social media.

Existing adversarial evaluation frameworks/benchmarks focus exclusively on model \textit{robustness}, typically requiring carefully controlled and \textit{expensive} human annotations~\cite{gui2021textflint,keila2021dynabench} that tend to result in small datasets;~e.g. Adversarial GLUE~\cite{wang2021adversarial} contains only 5,000 human-verified attacks. However, to the best of our knowledge, no dataset currently exists to support the task of attack identification. To address this issue, we propose TCAB, a benchmark dataset that comprises over 1.5 million fully-automated low-effort attacks, providing sufficient data to enable proper training and evaluation of models specific to the task~(and potential subtasks) of attack identification.

We summarize our contributions below, outlining the unique advantages of TCAB over existing adversarial benchmarks.
\begin{enumerate}
    
    \item We introduce \textit{attack identification} and its primary task \textit{attack labeling} -- automatically determining the adversarial attacks~(if any) used to generate a given piece of text. Attack identification also includes additional tasks such as \textit{attack detection}, \textit{attack target labeling}, \textit{attack localization}, and \textit{attack characterization}, enabling defenders to learn more about their attackers.

    \item We create TCAB, a text classification attack benchmark consisting of more than 1.5 million successful adversarial examples from a diverse set of twelve attack methods targeting three state-of-the-art text classification models trained on six sentiment/toxic-speech domain datasets. TCAB is designed to be expanded as new text attacks, classifiers, and domains are developed. This benchmark  supports research into attack identification and related tasks. 
    
    \item We adopt crowd-sourcing to evaluate a portion of TCAB and analyze the label-preserving nature of the attack methods. We find that 
    51\% and 81\% of adversarial instances preserve their original labels for sentiment and abuse datasets, respectively. 
    
    \item We present a baseline approach for attack detection and labeling that involves a combination of contextualized fine-tuned BERT embeddings and hand-crafted text, language model, and target model properties. Our baseline approach achieves 91.7\% and 66.7\% accuracy for attack detection and labeling, averaged over all datasets and target models.

\end{enumerate}

TCAB is available at \url{https://react-nlp.github.io/tcab/}, including the clean (source) instances, the manipulated attack instances, code for generating features and baseline models, and code for extending the dataset with new attacks and source datasets.


\section{Background and Problem Setup}
\label{sec:background}

Existing adversarial evaluation frameworks focus only on model \textit{robustness}. Robustness Gym~\cite{goel2021robustness_gym} and TextFLINT~\cite{gui2021textflint} allow users to measure the performance of their models on a variety of text transformations and adversaries.
Adversarial GLUE~\cite{wang2021adversarial} is a multi-task robustness benchmark that was created by applying 14 textual adversarial attack methods to GLUE tasks. Dynabench~\cite{keila2021dynabench} is a related framework for evaluating and training NLP models on adversarial examples created entirely by human adversaries.

Of these, TCAB is most similar to Adversarial GLUE. However, TCAB was designed for a different purpose --- attack identification rather than robustness evaluation. TCAB is also much larger than Adversarial GLUE~(1.5 million fully-automated attacks vs.~5,000 human-verified attacks), focuses only on classification, and includes multiple classification domain datasets.


In this work, we focus on text classifiers and attacks on them. Given an input sequence $\mathbf{x} = (x_1, x_2, ..., x_N) \in \mathcal{X}$~(the instance space), a text classifier $f$ maps $\mathbf{x}$ to a label $y \in \mathcal{Y}$, the set of output labels. For sentiment analysis, $\mathcal{Y}$ may be positive or negative sentiment; or for toxic comment detection, $\mathcal{Y}$ may be toxic or non-toxic.

A text-classification adversary aims to generate an adversarial example $\mathbf{x'}$ such that $f(\mathbf{x'}) \neq f(\mathbf{x})$. Ideally, the changes made on $\mathbf{x}$ to obtain $\mathbf{x'}$ are minimal such that a human would label them the same way. Perturbations may occur on the character-\cite{ebrahimi2018hotflip, eger2019viper}, token-\cite{gao2018deepwordbug}, word-\cite{jin2019textfooler}, or sentence-level~\cite{iyyer2018adversarial,wang2020t3}, or a combination of levels~\cite{li2018textbugger}; perturbations may also be structured such that certain input properties are preserved, such as the semantics~\cite{eger2019viper}, perplexity~\cite{alzantot2018genetic,jia2019certified}, fluency~\cite{ramakrishnan2020bae}, or grammar~\cite{zang2020pso}.

The primary task of attack identification is \textit{attack labeling}:~given a~(possibly) perturbed input sequence $\mathbf{x^*} \in M(\mathbf{x})$, in which $M(\mathbf{x})$ is a function that perturbs $\mathbf{x}$ using any one attack method from a set of attacks $S$ (including a ``clean'' attack in which the input is not perturbed), an attack labeler $f^{LAB}: \mathcal{X} \rightarrow S$ maps the perturbed sequence to an attack method in~$S$. Additionally, multiple subtasks complement the primary challenge of attack labeling; these include \textit{attack localization}, \textit{attack target labeling}, and \textit{attack characterization}~(see~\S\ref{sec:potential_tasks} for formal descriptions of all related tasks).

In pursuit of solving these problems, we develop and curate a large collection of adversarial attacks on a number of classifiers trained on various domain datasets. In the following section, we detail our process for generating this benchmark, describe its characteristics, and provide a human evaluation of the resulting dataset~(\S\ref{sec:tcab}). We then evaluate a set of baseline models on attack detection and labeling using our newly created benchmark~(\S\ref{sec:experiments}).

\begin{figure}[t]
  \centering
  \includegraphics[width=\textwidth]{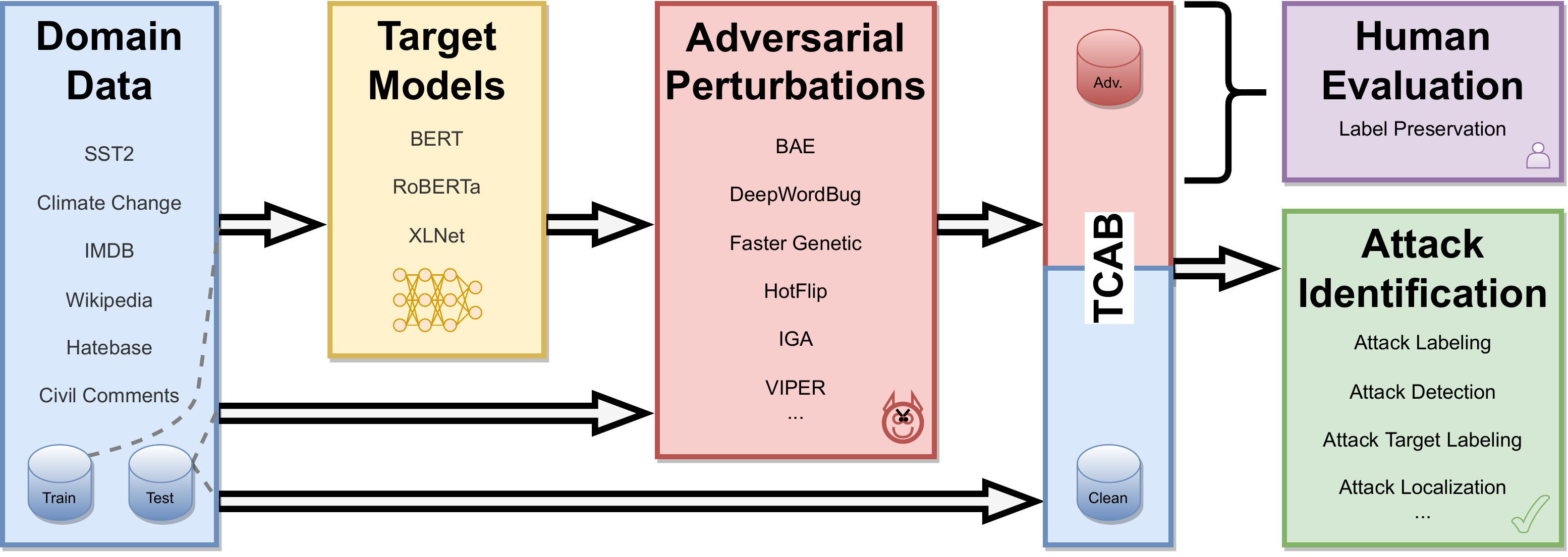}
  \caption{High-level overview of the TCAB generation and evaluation workflow.}
  \label{fig:tcab_overview}
\end{figure}

\section{TCAB Benchmark}
\label{sec:tcab}

We now present the Text Classification Attack Benchmark (TCAB), a dataset for developing and evaluating methods for identifying adversarial attacks against text classifiers. Here we describe the process for constructing this benchmark, and evaluate its characteristics.

\subsection{Domain Datasets}
\label{sec:domain_datasets}

Our focus is on two text-classification domains, sentiment analysis and abuse detection. Sentiment analysis is a popular and widely studied task~\cite{potts2021dynasent,yang2021exploring,zhong2021useradapter,shi2022effective,ling2022vision,wu2022adversarial} while abuse detection is more likely to be adversarial~\cite{pavlopoulos2020toxicity,zhang2020demographics,kennedy2020contextualizing}.

For sentiment analysis, we attack models trained on three domains: (1) \textbf{Climate Change}\footnote{\url{https://www.kaggle.com/edqian/twitter-climate-change-sentiment-dataset}}, 62,356 tweets on climate change; (2) \textbf{IMDB}~\cite{maas-EtAl:2011:ACL-HLT2011},~50,000 movie reviews, and (3) \textbf{SST-2}~\cite{socher2013recursive}, 68,221 movie reviews. For abuse detection, we attack models trained on three toxic-comment datasets: (1)~\textbf{Wikipedia} (Talk Pages)~\cite{wulczyn2017ex, dixon2018measuring}, 159,686 comments from Wikipedia administration webpages, (2) \textbf{Hatebase}~\cite{davidson2017automated}, 24,783 comments, and (3) \textbf{Civil Comments}\footnote{\url{https://www.kaggle.com/c/jigsaw-unintended-bias-in-toxicity-classification}}, 1,804,874 comments from independent news sites. All datasets are binary (positive vs.\ negative or toxic vs.\ non-toxic) except for Climate Change, which includes neutral sentiment. Additional dataset details are in the Appendix~\S\ref{app_sec:domain_datasets}.

\subsection{Target Models}

We finetune BERT~\cite{devlin2019bert}, RoBERTa~\cite{liu2019roberta}, and XLNet~\cite{yang2019xlnet} models — all from HuggingFace's transformers library~\cite{wolf2020transformers} — on the six domain datasets. We use transformer-based models since they represent current state-of-the-art approaches to text classification, and we use multiple architectures to obtain a wider range of adversarial examples, ultimately testing the robustness of attack identification models to attacks targeting different victim models.

Table~\ref{tab:victim_model_performance} shows the performance of these models on the test set of each domain dataset. On most datasets, RoBERTa slightly outperforms the other two models both in accuracy and AUROC. Training code and additional details such as selected hyperparameters are in the Appendix~\S\ref{app_sec:target_models}.

\subsection{Attack Methods}
\label{sec:attack_methods}

We select twelve different attack methods that cover a wide range of design choices and assumptions, such as model access level~(e.g., white/gray/black box), perturbation level~(e.g., char/word/token), and linguistic constraints.
Table~\ref{tab:attack_methods}~(Appendix,~\S\ref{app_sec:attack_samples}) provides a summary of all attack methods and their characteristics.

\paragraph{Target Model Access and Perturbation Levels.}

Of the twelve attack methods, only two~\cite{ebrahimi2018hotflip,li2018textbugger} have full access to the target model~(i.e., a white-box attack), while five~\cite{gao2018deepwordbug,jia2019certified,alzantot2018genetic,wang2019igawang,pruthi2019combating} assume some information about the target~(gray box), and the rest~\cite{ramakrishnan2020bae,zang2020pso,li2018textbugger,jin2019textfooler,eger2019viper} can only query the output~(black box). The majority of methods perturb entire words by swapping them with similar words based on sememes~\cite{zang2020pso}, synonyms~\cite{jin2019textfooler} or an embedding space~\cite{ramakrishnan2020bae,jia2019certified, alzantot2018genetic,wang2019igawang,pruthi2019combating}. The remaining methods~\cite{gao2018deepwordbug,ebrahimi2018hotflip,li2018textbugger,eger2019viper} operate on the token/character level, perturbing the input by inserting/deleting/swapping different characters.

\paragraph{Linguistic Constraints.}

Linguistic constraints promote indistinguishable attacks. For example, Genetic~\cite{alzantot2018genetic}, FasterGenetic~\cite{jia2019certified}, HotFlip~\cite{ebrahimi2018hotflip}, and Pruthi~\cite{pruthi2019combating} limit the number or percentage of words perturbed. Other methods ensure the distance between the perturbed text and the original text is ``close'' in some embedding space; for example, BAE~\cite{ramakrishnan2020bae}, TextBugger~\cite{li2018textbugger}, and TextFooler~\cite{jin2019textfooler} constrain the perturbed text to have high cosine similarity to the original text using a universal sentence encoder~(USE)~\cite{cer2018universal}, while IGA~\cite{wang2019igawang} and VIPER~\cite{eger2019viper} ensure similarity in word and visual embedding spaces, respectively. Some methods, such as TextBugger and TextFooler, use a combination of constraints to further limit deviations from the original input.

\paragraph{Attack Toolchains.} We use TextAttack~\cite{morris2020textattack} and OpenAttack~\cite{zeng2020openattack} --- open-source toolchains that provide fully-automated off-the-shelf attacks --- to generate adversarial examples. For these toolchains, attack methods are implemented using different search methods. For example, BAE~\cite{ramakrishnan2020bae}, DeepWordBug~\cite{gao2018deepwordbug}, TextBugger~\cite{li2018textbugger}, and TextFooler~\cite{jin2019textfooler} use a word importance ranking to greedily decide which word(s) to perturb for each query; in contrast, Genetic~\cite{alzantot2018genetic} and PSO~\cite{zang2020pso} use a genetic algorithm and particle swarm optimization to identify word-perturbation candidates, respectively. For all attack methods, we set a maximum limit of 500 queries per instance. Note an attack method may be implemented by \textit{both} toolchains~(e.g., TextBugger is implemented by TextAttack and TextBugger).

\begin{table}[t]
\caption{Predictive performance of the target models on the test set for each domain dataset; *: multiclass-macro-averaged AUC; the rest are binary-classification tasks.}
\center
\begin{tabular}{l ll ll ll}
\toprule
 & \multicolumn{2}{c}{\textbf{BERT}}
 & \multicolumn{2}{c}{\textbf{RoBERTa}}
 & \multicolumn{2}{c}{\textbf{XLNet}}
\\
 \cmidrule(lr){2-3}
 \cmidrule(lr){4-5}
 \cmidrule(lr){6-7}
 \textbf{Dataset} & Acc. & AUC & Acc. & AUC & Acc. & AUC \\
\midrule
Climate Change*    & 79.8 & 0.899 & \textbf{81.2} & \textbf{0.917} & 80.1 & 0.910 \\
IMDB               & 87.0 & 0.949 & \textbf{90.7} & \textbf{0.968} & 90.1 & 0.965 \\
SST-2                & 91.8 & 0.972 & \textbf{92.7} & \textbf{0.978} & 92.3 & 0.974 \\
Wikipedia          & 96.5 & 0.982 & \textbf{96.6} & \textbf{0.985} & 96.4 & 0.983 \\
Hatebase           & \textbf{95.8} & 0.983 & \textbf{95.8} & \textbf{0.987} & 93.9 & 0.979 \\
Civil Comments     & \textbf{95.2} & \textbf{0.968} & 95.1 & 0.967 & 95.0 & 0.965 \\
\bottomrule
\end{tabular}
\label{tab:victim_model_performance}
\end{table}

\subsection{Dataset Generation}
\label{sec:dataset_generation}

To create TCAB, we perturb examples from the test sets of the six domain datasets~(see \S\ref{sec:domain_datasets} for domain datasets). SST-2, Wikipedia, and IMDB have predefined train/test splits. For the other three datasets, we randomly partition the data into an 80/10/10 train/validation/test split.

For each model/domain dataset combination, we only attack test set examples in which the model's prediction is correct. For abuse datasets, we further constrain our focus on examples in the test set that are both predicted correctly \emph{and} toxic; perturbing non-toxic text to be classified as toxic is a less likely adversarial task.\footnote{One can imagine cases where an attacker would have this goal, such as trying to get someone else banned from a social network by tricking them into posting text that triggers an abuse filter. However, we expect this to be much less common than attackers simply trying to evade abuse filters.}

The full pipeline for generating TCAB is shown in Figure~\ref{fig:tcab_overview}.

\paragraph{Adding Clean Instances and Creating Development-Test Splits.} For each dataset, we randomly sample a fraction of the instances from the test set to include as ``clean'' unperturbed examples. After merging the clean and adversarial examples, we split the data into a 60/20/20 train/validation/test split. Additionally, all instances with the same \textit{source text} are sent to the same split (i.e., multiple successful attacks on the same original text are all sent to the same split) to avoid any data leakage. The train and validation sets are publicly available,\footnote{\url{https://zenodo.org/record/7226519}} while the test set is available upon request.

\paragraph{Extending TCAB.}
\label{sec:extending_tcab}

TCAB is designed to incorporate new attacks as they develop, or existing attacks on new text classifiers. TCAB thus facilitates research into attack identification models that stay up-to-date with the latest attacks on the latest text classifiers. Instructions and code for extending TCAB with new attacks or domain datasets is available at~\url{https://github.com/REACT-NLP/tcab_generation}.

\subsection{TCAB Statistics}

TCAB contains a total of 1,504,607 successful attacks on six domain datasets against three different target models using twelve attack methods from two open-source toolchains.

\begin{figure}[t]
  \centering
  \includegraphics[width=\textwidth]{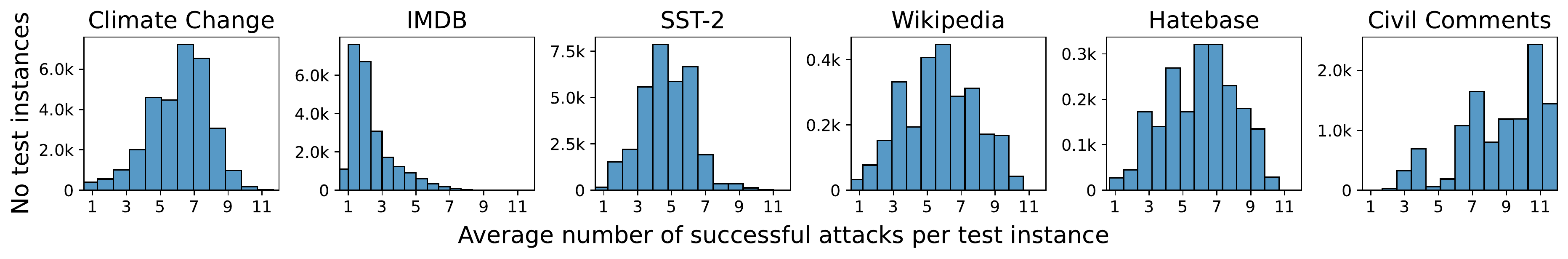}
  \caption{Histogram of the number of successful attacks~(out of 12) averaged across all three target models for each domain dataset.}
  \label{fig:success_dists}
\end{figure}

\paragraph{Attack Success Rate.} Table \ref{tab:success-rates} shows a breakdown of attack success rates and the number of successful attacks for each method. For Civil Comments, many instances were very easy to manipulate successfully~(Figure \ref{fig:success_dists}: far right), and it was not uncommon for all 12 attackers to successfully perturb the same instance. In contrast, it was quite rare for more than three of the attackers to be successful on any IMDB instance. Of the three target-model architectures, XLNet was the most robust — it was successfully attacked 61\% of the time (this percentage is computed over all attack attempts made against all XLNet models). BERT and RoBERTa were fooled 63\% and 66\% of the time, respectively.

\begin{figure}[ht]
  \centering
  \includegraphics[width=0.495\textwidth]{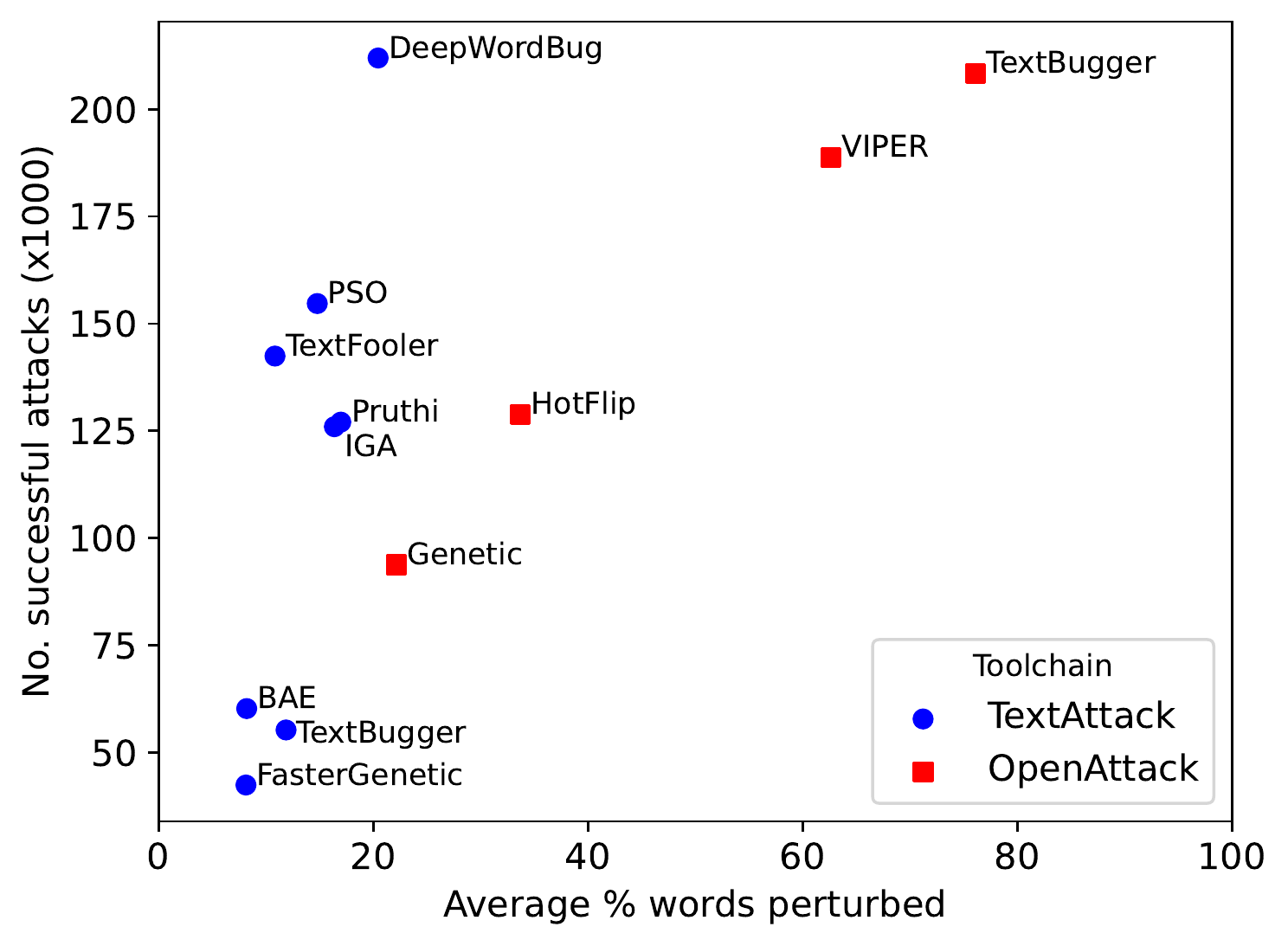}
  \includegraphics[width=0.495\textwidth]{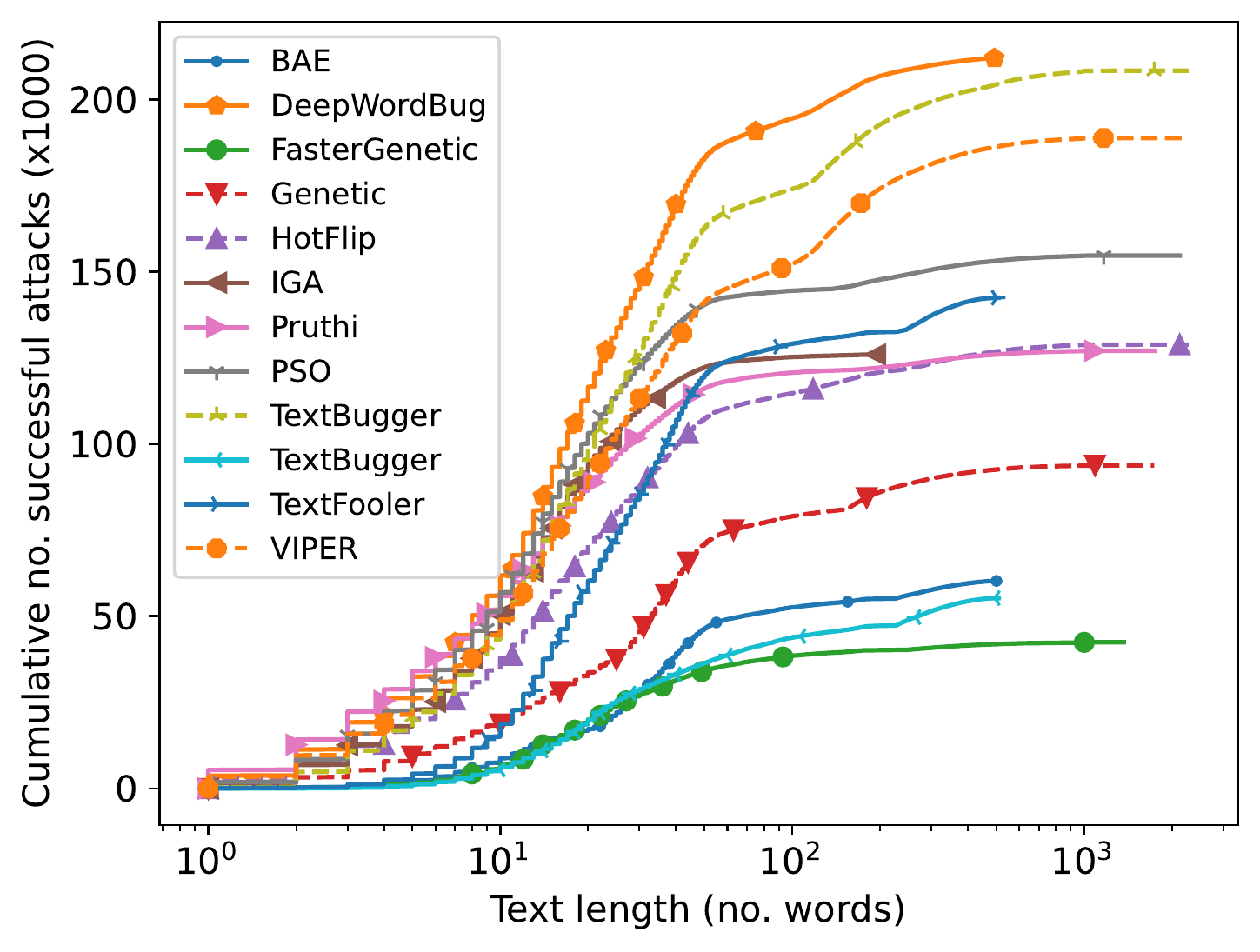}
  \caption{\textit{Left}:~Average percentage of words perturbed per successful attack vs. the number of successful attacks across all domains, with TextAttack as blue circles and OpenAttack as red squares. \textit{Right}:~Cumulative number of successful attacks across all domains as the size of the input text increases, with TextAttack and OpenAttack shown as solid and dashed lines, respectively.}
  \label{fig:num_success}
\end{figure}

\paragraph{Perturbation Size.} Interestingly, the degree of input perturbation and attack-success frequency are only weakly correlated~(Figure~\ref{fig:num_success}:~left). For example, DeepWordBug perturbs just 20\% of the input words on average, but generates more successful attacks than any other method. The four attack methods from OpenAttack perturb more words in the input than any of the TextAttack methods.

\paragraph{Text Length.} Figure~\ref{fig:num_success}~(right) shows the cumulative number of successful attacks for each attack method as the length of the input text increases~(across all datasets). Overall, FasterGenetic, TextBugger, and BAE were less effective than other methods on short-($<10$ words) and medium-($10-100$ words) sized inputs. DeepWordBug successfully attacked most instances~$<500$ words; however, DeepWordBug, TextBugger, BAE, TextFooler, and especially IGA failed to successfully attack very long inputs~($>1000$ words).

\begin{table}[ht]
\center
\caption{Percentage (and number) of successful attacks across all three target models. Attack methods with an ``*'' are from the OpenAttack toolchain, those without are from the TextAttack toolchain.}
\label{tab:success-rates}
\setlength{\tabcolsep}{5pt}
\small
\begin{tabular}{lcccccc}
\toprule
\textbf{Attack Method} & \textbf{Clim. Cha.} & \textbf{IMDB} & \textbf{SST-2} & \textbf{Wikipedia} & \textbf{Hatebase} & \textbf{Civ. Com.} \\ 
\midrule
\textbf{BAE}~\cite{ramakrishnan2020bae} &	52 (21.4k)  &  36 (\hphantom{0}5.1k)  &68 (\hphantom{0}4.0k)  &61 (4.5k)  &61 (3.6k)  &71 (21.6k)   \\
DeepWordBug~(\textbf{DWB})~\cite{gao2018deepwordbug} &	86 (80.9k)  &  74 (16.6k)  &  79 (74.1k)  &  79 (5.8k)  &76 (4.6k)  &99 (30.0k)   \\
FasterGenetic~(\textbf{FG})~\cite{jia2019certified} &	38 (14.1k)  &  11 (\hphantom{0}2.2k)  &30 (\hphantom{0}1.8k)  &32 (2.4k)  &33 (2.0k)  &66 (19.9k)   \\
Genetic*~(\textbf{Gn.*})~\cite{alzantot2018genetic} &	67 (24.2k)  &  46 (11.7k)  &  34 (29.5k)  &  45 (3.4k)  &13 (0.8k)  &80 (24.3k)  \\
HotFlip*~(\textbf{HF*})~\cite{ebrahimi2018hotflip} &	52   (49.5k)  &  36 (12.3k)  &  42 (39.5k)  &  37 (2.8k)  &35 (2.1k)  &75 (22.6k)  \\
\textbf{IGA}~\cite{wang2019igawang}  &	52 (49.1k)  &  \hphantom{0}0 (\hphantom{0.0k}0) & 59 (54.3k)  &  \hphantom{0}0 (\hphantom{0.k}0) & 54 (3.7k)  & 64 (19.3k) \\
Pruthi~(\textbf{Pr.})~\cite{pruthi2019combating} &	43 (40.7k)  &  19 (\hphantom{0}5.2k)  &59 (55.9k)  &  35 (2.6k)  &40 (2.4k)  &67 (20.3k)  \\
\textbf{PSO}~\cite{zang2020pso} &	59 (55.7k)  &  27 (\hphantom{0}9.3k)  &72 (66.7k)  &  31 (2.3k)  &35 (2.1k)  &62 (18.7k)  \\
TextBugger*~(\textbf{TB*}) \cite{li2018textbugger} &	81 (75.9k)  &  79 (33.1k)  &  65 (61.4k)  &  74 (5.5k)  &54 (3.2k)  &97 (29.3k)  \\
TextBugger~(\textbf{TB})~\cite{li2018textbugger} &	74 (\hphantom{0}6.3k)  &57 (\hphantom{0}8.0k)  &68 (\hphantom{0}4.0k)  &65 (4.8k)  &56 (3.4k)  &95 (28.8k)  \\
TextFooler~(\textbf{TF})~\cite{jin2019textfooler} &	92 (86.5k)  &  51 (\hphantom{0}9.9k)  &94 (\hphantom{0}5.5k)  &82 (6.0k)  &83 (5.0k)  &98 (29.5k)  \\
VIPER*~(\textbf{VIP*})~\cite{eger2019viper} &	62 (58.7k)  &  88 (38.8k)  &  63 (59.6k)  &  66 (4.9k)  &67 (4.0k)  &75 (22.8k)  \\
\bottomrule
\end{tabular}
\end{table}

\begin{table}[ht]
\center
\caption{Percentage of adversarial instances whose labels were preserved after human evaluation; each instance was labeled by five different workers from Amazon Mechanical Turk.}
\label{tab:human_eval_datasets}
\begin{tabular}{cccccc}
\toprule
\textbf{Climate Change} & \textbf{IMDB} & \textbf{SST-2} & \textbf{Wikipedia} & \textbf{Hatebase} & \textbf{Civil Comments} \\
\midrule
    48.3 & 52.4 & 50.7 & 81.2 & 80.7 & 81.3 \\
\bottomrule
\end{tabular}
\end{table}

\subsection{Human Evaluation}

Text attacks try to change the predicted labels, but often change the meaning of the text as well~\cite{morris-2020-reevaluating}. We have also observed this; for example, in Table~\ref{tab:attack_samples_brief} for example, IGA changes the word ``amateurish'' to ``enthusiastic,'' which changes the sentiment (and thus the ``true'' label) from negative to positive. To estimate how often the original label is preserved after an attack in TCAB,
we conduct a human evaluation on a portion of the perturbed examples using Amazon Mechanical Turk (MTurk)~\footnote{Amazon Mechanical Turk, \url{https://www.mturk.com/}.}.


First, we randomly selected 30 adversarial examples for each attack method/target model/domain dataset combination,\footnote{For IMDB, we only select 15 since these texts are much longer on average.} and presented workers with the text for one of these adversarial examples at a time. We then asked workers to evaluate the sentiment/toxicity of the adversarial text on a scale from $1$ to $5$ where $5$ indicates \textit{strongly positive} for sentiment analysis and \textit{strongly toxic} for abuse detection. Qualification requirements, compensation, and user interface details are in the Appendix,~\S\ref{app_sec:human_eval}.
Overall, 5,581 adversarial instances received labels from five different workers. We marked each instance's label as preserved if the median score of the workers was above 3 for examples from the abuse domain, and below or above 3 for examples with negative and positive sentiment for the sentiment analysis domain, respectively\footnote{The Climate Change dataset has 3 sentiments: negative, neutral, and positive; for neutral-labeled instances from this dataset, we mark the label as preserved if the median score of the annotators is equal to 3.}

\paragraph{Results.}

Across all instances evaluated, 67.6\% of the labels were preserved. However, instances from abuse datasets were significantly more likely to be preserved than instances from sentiment analysis datasets; Table~\ref{tab:human_eval_datasets} shows the percentage of labels preserved for abuse datasets ranges from 80.7--81.3\% while the percentage for sentiment analysis datasets ranges from 48.3--52.4\%. For target models, the percentages of labels preserved~(over all datasets) is 68.9\%, 65.4\%, and 68.6\% for BERT, RoBERTa, and XLNet, respectively. For different attack methods, the percentage of labels preserved ranges from 61.1--73.9\%~(see~\S\ref{app_sec:human_eval}, Table~\ref{tab:human_eval_attacks} for exact values for each attack). 

Note our human evaluation focuses only on label preservation, a narrow aspect of total semantic preservation, which may include analyzing how much the perturbed text differs from the original text, often resulting in significantly lower preservation rates as shown in~\citet{morris-2020-reevaluating}.



\subsection{Potential Tasks}
\label{sec:potential_tasks}

We provide a large collection of fully-automated adversarial attacks on text classifiers. The dataset presents several challenges to the adversarial NLP community.

\paragraph{Primary Task: Attack Detection.}

\textit{Attack detection} determines if any perturbation is present on a given piece of text~\cite{zhou2019learning,pruthi2019combating,mozes-etal-2021-frequency,le2020detecting,wallace2019universal,hovy2016enemy,Li2021ContextualizedPF}. 
Formally, given $\mathbf{x^*} \in \mathcal{X}$, a model $f^{DET}: \mathcal{X} \rightarrow \{-1, +1\}$ detects the presence of any perturbations on $\mathbf{x^*}$. 
Accurately predicting the presence of an attack complements
the much more difficult attack labeling task.

\paragraph{Primary Task: Attack Labeling.}

The main challenge of TCAB is \textit{attack labeling} --- classifying which attack~(if any) perturbed a given piece of text; accurate predictions on this task can provide more information about the attack/attacker. 
Formally, given $\mathbf{x^*} \in \mathcal{X}$, where $\mathbf{x^*} = M(\mathbf{x})$ for some (unknown) perturbation function $M \in S$ and clean instance $\mathbf{x}$,
an attack labeler $f^{LAB}: \mathcal{X} \rightarrow S$ maps the~(possibly) perturbed sequence to an attack method in~$S$.
$S$ is assumed to be known, finite, and include a ``clean'' attack in which the input is not perturbed.

\paragraph{Subtask: Attack Localization.}

\textit{Attack localization} identifies which~(if any) tokens/chars/words are perturbed in a given piece of text. Knowing the location of an attack provides information about what words the attacker is likely to target, and allows efficient reparation of the perturbed instance. 
More formally, given $\mathbf{x^*} \in \mathcal{X}$, a model $f^{LOC}: \mathcal{X} \rightarrow (b_1, b_2, \ldots, b_N)$ identifies the perturbed tokens in~$\mathbf{x^*}$, where~$b_i$ is a boolean in which~$1$ means~$x_i$ is perturbed, and~$0$ means~$x_i$ is not perturbed.

\paragraph{Subtask: Attack Target Labeling.}

Attack methods aimed at one target model may result in significantly different perturbations when applied to a different target model~(e.g., RoBERTa vs. XLNet). Thus, \textit{attack target labeling} aims to identify the target model being attacked. More formally, given $\mathbf{x^*} \in \mathcal{X}$, a model $f^{TML}: \mathcal{X} \rightarrow \mathcal{T}$ maps the~(possibly) perturbed input sequence to a model in a set of possible target models~$\mathcal{T}$.

\paragraph{Subtask: Attack Characterization.}

As described in~\S\ref{sec:attack_methods}, attack methods make different design choices and assumptions about model access/perturbation levels and linguistic constraints. \textit{Attack characterization} comprises multiple subtasks to
predict these properties from an attack instance:
\textit{Access Level} (white box, gray box, or black box), \textit{Perturbation Level} (token, char, word, or some combination), and \textit{Linguistic Constraints} (if any).

%
%
Initial work in this direction by \citet{xie-etal-2021-models} has shown example usage of these tasks and how they can extract additional information from attackers.



\section{Baseline Models}
\label{sec:experiments}

\begin{table}[t]
\small
\setlength\tabcolsep{3pt}
\center
\caption{Attack detection and labeling results showing the balanced accuracy of each baseline model on each dataset for attacks targeting RoBERTa.}
\label{tab:baselines_roberta}
\begin{tabular}{l ccc c ccc}
\toprule
\multirow{2}{*}{}
  & \multicolumn{3}{c}{\textbf{Attack Detection} (2 classes)}
  &
  & \multicolumn{3}{c}{\textbf{Attack Labeling} (12 classes)} \\
\cmidrule(lr){2-4}\cmidrule(lr){6-8}
 \textbf{Dataset} &
 FT &
 LR\scriptsize-FT-TLC &
 LGB\scriptsize-FT-TLC &
 & FT &
 LR\scriptsize-FT-TLC &
 LGB\scriptsize-FT-TLC \\
\midrule
Climate Change
 & 91.1 & 91.6 & \textbf{91.8}
 &
 & 77.5 & 80.0 & \textbf{80.4}
\\
IMDB
 & 93.9 & 95.7 & \textbf{97.3}
 &
 & 69.3 & 73.8 & \textbf{78.9}
\\
SST-2
 & \textbf{92.8} & 90.4 & 92.0
 &
 & 59.0 & 64.7 & \textbf{65.8}
\vspace{0.01cm} \\
\hdashline \\
\vspace{-0.6cm} \\
Wikipedia
 & 88.2 & 91.4 & \textbf{92.8}
 &
 & 46.2 & 48.9 & \textbf{56.7}
\\
Hatebase
 & 90.2 & \textbf{93.3} & \textbf{93.3}
 &
 & 58.2 & 63.6 & \textbf{64.0}
\\
Civil Comments
 & 87.1 & 89.0 & \textbf{90.5}
 &
 & 60.0 & 62.4 & \textbf{64.2}
\\
    \bottomrule
    \end{tabular}
\end{table}

Here we present a baseline approach for tackling the attack detection and labeling tasks, and evaluate our approach on the TCAB benchmark.

\paragraph{Fine-Tuned BERT, FT.}

As an initial baseline, we fine-tune a pretrained BERT~\cite{devlin2019bert} model to automatically engineer features and directly perform attack detection or labeling.

\paragraph{Fine-Tuned BERT + Hand-Crafted Features, FT-TLC.}

In addition to fine-tuned BERT, we engineer a set of hand-crafted features that represent different properties of the input, namely text, language model, and target model properties. Text properties~(\textbf{T}) represent surface-level characteristics of the input such as the length, non-ASCII character count, token casing/shape, punctuation marks, and more~(see~\S\ref{app_sec:engineered_features} for a complete list). Language model properties~(\textbf{L}) identify structural parts of the text, such as ungrammatical, awkward, or generic phrasing using a pretrained language model~(e.g., BERT~\cite{devlin2019bert}, GPT-3~\cite{brown2020language}). Target model properties~(\textbf{C}) use the target model's output posteriors, node activations, gradients, and saliency to capture any changes in the target model due to deceptive input; this measures the effect of deceptive text on the target classifier. We use the same fine-tuned BERT to generate contextualized embeddings of the input, and concatenate the output to the full set of hand-crafted features we extract from the input, creating a fixed-length feature vector. We then train a logistic regression model~(\textbf{LR}) or gradient-boosted tree ensemble~(\textbf{LGB}~\cite{ke2017lightgbm}) to perform attack detection or labeling. Figure~\ref{fig:experiment_pipeline} in~\S\ref{app_sec:baseline_models} provides an overview of this combined approach.

\paragraph{Setup and Results.}

For both tasks, we balance all classes via oversampling. We perform attack detection as a binary classification task with 2 classes~(clean and perturbed) and attack labeling as a multiclass classification task with 12 classes~(11 attacks + clean).\footnote{Since TextBugger was implemented by both TextAttack and OpenAttack, we merge their instances here to consider them as a single attack, resulting in 11 attacks total.} Table~\ref{tab:baselines_roberta} shows attack detection and labeling results for attacks against the RoBERTa target model. Attack detection accuracy of the best model on each dataset ranges from 90.5-97.3\%; detection accuracy is similar for the other target models~(\S\ref{app_sec:baseline_results}, Tables~\ref{tab:baselines_bert}~and~\ref{tab:baselines_xlnet}), with an average of 91.9\%, 93.2\%, and 91.0\% for attacks targeting BERT, RoBERTa, and XLNet, respectively. For attack labeling, accuracy for the best model on each dataset ranges from 56.7-80.4\%~(note a random model would get~$\sim8.33$\%), with average accuracies of 64.2\%, 68.1\%, and 67.6\% for attacks targeting BERT, RoBERTa, and XLNet, respectively.

Overall, combining fine-tuned BERT with hand-crafted features works best in all cases for attack labeling and all but one case for attack detection. Fine-tuned BERT achieves decent predictive performance on its own; however, we find adding additional features, especially target model features, consistently improves performance for these tasks, with LGB slightly outperforming LR.


\section{Conclusion}

We introduce TCAB, a benchmark for analyzing, understanding, detecting, and labeling adversarial attacks on text classifiers. We attack three different text classifiers trained on six domain datasets using twelve attack methods to obtain over 1.5 million successful adversarial attacks, and evaluate the label-preserving qualities of these attacks with human annotations. TCAB is easily extendable, facilitating research into attack identification models that stay up-to-date with the latest attacks.

\begin{ack}
This work was supported by the Defense Advanced Research Projects Agency (DARPA), agreement number HR00112090135. This work benefited from access to the University of Oregon high performance computer, Talapas.
\end{ack}

\bibliographystyle{plainnat}
\bibliography{main}

\section*{Checklist}


\begin{enumerate}

\item For all authors...
\begin{enumerate}
  \item Do the main claims made in the abstract and introduction accurately reflect the paper's contributions and scope?
    \answerYes{}
  \item Did you describe the limitations of your work?
    \answerYes{See~\S\ref{app_sec:limitations}.}
  \item Did you discuss any potential negative societal impacts of your work?
    \answerYes{See~\S\ref{app_sec:impact}.}
  \item Have you read the ethics review guidelines and ensured that your paper conforms to them?
    \answerYes{}
\end{enumerate}

\item If you are including theoretical results...
\begin{enumerate}
  \item Did you state the full set of assumptions of all theoretical results?
    \answerNA{}
	\item Did you include complete proofs of all theoretical results?
    \answerNA{}
\end{enumerate}

\item If you ran experiments (e.g. for benchmarks)...
\begin{enumerate}
  \item Did you include the code, data, and instructions needed to reproduce the main experimental results (either in the supplemental material or as a URL)?
    \answerYes{See~\S\ref{app_sec:tcab_details} and~\S\ref{app_sec:baseline_models}.}
  \item Did you specify all the training details (e.g., data splits, hyperparameters, how they were chosen)?
    \answerYes{See~\S\ref{app_sec:tcab_details}}.
	\item Did you report error bars (e.g., with respect to the random seed after running experiments multiple times)?
    \answerNo{}
	\item Did you include the total amount of compute and the type of resources used (e.g., type of GPUs, internal cluster, or cloud provider)?
    \answerYes{See~\S\ref{app_sec:tcab_details}}.
\end{enumerate}

\item If you are using existing assets (e.g., code, data, models) or curating/releasing new assets...
\begin{enumerate}
  \item If your work uses existing assets, did you cite the creators?
    \answerYes{See~\S\ref{sec:domain_datasets}}.
  \item Did you mention the license of the assets?
    \answerYes{See~\S\ref{app_sec:dataset_documentation}.}
  \item Did you include any new assets either in the supplemental material or as a URL?
    \answerYes{See~\url{https://react-nlp.github.io/tcab/} for details.}
  \item Did you discuss whether and how consent was obtained from people whose data you're using/curating?
    \answerNA{}
  \item Did you discuss whether the data you are using/curating contains personally identifiable information or offensive content?
    \answerYes{See~\S\ref{app_sec:dataset_documentation}.}
\end{enumerate}

\item If you used crowdsourcing or conducted research with human subjects...
\begin{enumerate}
  \item Did you include the full text of instructions given to participants and screenshots, if applicable?
    \answerYes{See~\S\ref{app_sec:UI}.}
  \item Did you describe any potential participant risks, with links to Institutional Review Board (IRB) approvals, if applicable?
    \answerNA{}
  \item Did you include the estimated hourly wage paid to participants and the total amount spent on participant compensation?
    \answerYes{See~\S\ref{app_sec:compensation}}
\end{enumerate}

\end{enumerate}


\appendix

\newpage
\section{Dataset Documentation}
\label{app_sec:dataset_documentation}

The TCAB benchmark dataset is publicly available online at \url{https://react-nlp.github.io/tcab/}. This dataset has a CC-BY 4.0 license and its DOI is \textbf{10.5281/zenodo.7226519}. The dataset is represented in an easily read CSV format. Code for generating this dataset is available at \url{https://github.com/REACT-NLP/tcab_generation}.

\subsection{Datasheet}

These questions are copied from ``Datasheets for Datasets''~\cite{gebru2021datasheets}.

\subsubsection*{Motivation}

\begin{itemize}
    \item \textbf{For what purpose was the dataset created?} \eat{Was there a specific task in mind? Was there a specific gap that needed to be filled? Please provide a description.}
    
    We created TCAB to facilitate future research on analyzing, understanding, detecting, and labeling adversarial attacks for text classifiers.
    
    \item \textbf{Who created the dataset (e.g., which team, research group) and on behalf of which entity (e.g., company, institution, organization)?} 
    
    TCAB was created by the REACT-NLP team, consisting of researchers at the University of Oregon and University of California Irvine. The project was led by Daniel Lowd and Sameer Singh, and the dataset was constructed by Kalyani Asthana, Zhouhang Xie, Wencong You, Adam Noack, and Jonathan Brophy.

    \item \textbf{Who funded the creation of the dataset?} \eat{If there is an associated grant, please provide the name of the grantor and the grant name and number.}

    This work was supported by the Defense Advanced Research Projects Agency (DARPA), agreement number HR00112090135.

\end{itemize}

\subsubsection*{Composition}

\begin{itemize}
    \item \textbf{What do the instances that comprise the dataset represent (e.g., documents, photos, people, countries)?} \eat{Are there multiple types of instances (e.g., movies, users, and ratings; people and interactions between them; nodes and edges)? Please provide a description.}
    
    The dataset contains text only. 
    
    \item \textbf{How many instances are there in total (of each type, if appropriate)?}
    
    In total, TCAB contains 2,414,594 instances:~1,504,607 successful adversarial instances, and~909,987 ``clean'' unperturbed examples. See Table~\ref{tab:success-rates} for detailed statistics for adversarial instances by domain dataset and attacking algorithms. 

    \item \textbf{Does the dataset contain all possible instances or is it a sample (not necessarily random) of instances from a larger set?} \eat{If the dataset is a sample, then what is the larger set? Is the sample representative of the larger set (e.g., geographic coverage)? If so, please describe how this representativeness was validated/verified. If it is not representative of the larger set, please describe why not (e.g., to cover a more diverse range of instances, because instances were withheld or unavailable).}
    
    We attack the test sets of each domain dataset, thus adversarial examples in TCAB comprise attack instances from a portion of the original domain datasets.
    

    \item\textbf{What data does each instance consist of? \eat{“Raw” data (e.g., unprocessed text or images) or features?} In either case, please provide a description.}
    
    Each instance contains the original text, perturbed text, original ground-truth label, and whether the instance is ``clean'' or perturbed. For perturbed instances, we also provide information about the attack such as the method used, toolchain, maximum number of queries, time taken per attack, and more.  Please refer to our dataset page~(\url{https://zenodo.org/record/7226519}) for a full description of all attributes.

    \item \textbf{Is there a label or target associated with each instance?} \eat{If so, please provide a description.}
    
    We provide labels indicating (1) whether the text is an adversarial instance and (2) what algorithm created the corresponding adversarial instance, which could then be used as labels for detecting adversarial attacks and identifying the attacking algorithm.

    \item \textbf{Is any information missing from individual instances?} \eat{If so, please provide a description, explaining why this information is missing (e.g., because it was unavailable). This does not include intentionally removed information, but might include, e.g., redacted text.}
    
    There is no missing information from our instances.

    \item \textbf{Are relationships between individual instances made explicit (e.g., users' movie ratings, social network links)?} \eat{If so, please describe how these relationships are made explicit.}
    
    There are no explicit relationships between instances. However, our metadata supports grouping operations such as collecting instances perturbed from the same original text and instances attacked by the same attacking algorithm.

    \item \textbf{Are there recommended data splits (e.g., training, development/validation, testing)?} \eat{If so, please provide a description of these splits, explaining the rationale behind them.}
    
    We randomly split the original dataset into training, validation and test sets. To avoid potential information leakage, all instances that originate from the same original text (including the unperturbed text) are guaranteed to be partitioned to the same split.

    \item \textbf{Are there any errors, sources of noise, or redundancies in the dataset?} \eat{If so, please provide a description.}
    
    Yes, adversarial attacks are by nature noisy, and there are cases where human judges believe the attack flips the \textit{true} label of the text being perturbed. We present human evaluation results to help researchers evaluate the degree of such scenario in our dataset.
    
    \item \textbf{Is the dataset self-contained, or does it link to or otherwise rely on external resources (e.g., websites, tweets, other datasets)?} \eat{If it links to or relies on external resources, a) are there guarantees that they will exist, and remain constant, over time; b) are there official archival versions of the complete dataset (i.e., including the external resources as they existed at the time the dataset was created); c) are there any restrictions (e.g., licenses, fees) associated with any of the external resources that might apply to a future user? Please provide descriptions of all external resources and any restrictions associated with them, as well as links or other access points, as appropriate.} 
    
    The dataset is self-contained and does not have external dependencies.

    \item \textbf{Does the dataset contain data that might be considered confidential (e.g., data that is protected by legal privilege or by doctor–patient confidentiality, data that includes the content of individuals’ non-public communications)?} \eat{If so, please provide a description.}
    
    The dataset does not contain confidential information.

    \item \textbf{Does the dataset contain data that, if viewed directly, might be offensive, insulting, threatening, or might otherwise cause anxiety?} \eat{If so, please describe why.}
    
    Some instances from our dataset are derived from hatespeech datasets, and thus may contain offensive content.

    \item \textbf{Does the dataset relate to people?} \eat{If not, you may skip the remaining questions in this section.}
    
    The dataset is not related to people.
\end{itemize}

\subsubsection*{Collection Process}

\begin{itemize}
    \item \textbf{How was the data associated with each instance acquired?} \eat{Was the data directly observable (e.g., raw text, movie ratings), reported by subjects (e.g., survey responses), or indirectly inferred/derived from other data (e.g., part-of-speech tags, model-based guesses for age or language)? If data was reported by subjects or indirectly inferred/derived from other data, was the data validated/verified? If so, please describe how.}
    
    Our dataset is derived from publically available datasets. See section~\ref{app_sec:domain_datasets} for details.

    \item \textbf{What mechanisms or procedures were used to collect the data (e.g., hardware apparatus or sensor, manual human curation, software program, software API)?} How were these mechanisms or procedures validated?
    
    We use OpenAttack~\cite{zeng2020openattack} and TextAttack~\cite{morris2020textattack} --- open-source toolchains that provide fully-automated off-the-shelf attacks --- to generate adversarial examples for TCAB. We ensure all released adversarial instances successfully flips the prediction of the victim model. We conduct additional human evaluation to check the quality of generated adversarial instances.

    \item \textbf{If the dataset is a sample from a larger set, what was the sampling strategy (e.g., deterministic, probabilistic with specific sampling probabilities)?}
    
    We randomly split the data into a train/validation/test split for each domain dataset. We then train different target models using the training and validation sets, and generate adversarial instances by attacking the target model on the test set.

    \item \textbf{Who was involved in the data collection process (e.g., students, crowdworkers, contractors) and how were they compensated (e.g., how much were crowdworkers paid)?}
    
    Our data collection process is automated. First, we train multiple target model classifiers for each domain dataset, then we attack those target models using publicly-available attack algorithms to generate adversarial instances. See section~\ref{sec:dataset_generation} for detailed descriptions. 

    \item \textbf{Over what timeframe was the data collected?} \eat{Does this timeframe match the creation timeframe of the data associated with the instances (e.g., recent crawl of old news articles)? If not, please describe the timeframe in which the data associated with the instances was created.}
    
    We generated attacks and curated instances for TCAB over a period of 12 months, from November 1, 2020 to November 1, 2021.

    \item \textbf{Were any ethical review processes conducted (e.g., by an institutional review board)?} \eat{If so, please provide a description of these review processes, including the outcomes, as well as a link or other access point to any supporting documentation.}
    
    No.

    \item \textbf{Does the dataset relate to people?} \eat{If not, you may skip the remaining questions in this section.}
    
    The dataset is not related to people.
\end{itemize}

\subsubsection*{Preprocessing/cleaning/labeling}

\begin{itemize}

    \item \textbf{Was any preprocessing/cleaning/labeling of the data done (e.g., discretization or bucketing, tokenization, part-of-speech tagging, SIFT feature extraction, removal of instances, processing of missing values)?} \eat{If so, please provide a description. If not, you may skip the remainder of the questions in this section.}
    
    No, we directly use the text instances from each domain dataset.

    


\end{itemize}

\subsubsection*{Uses}

\begin{itemize}

    \item \textbf{Has the dataset been used for any tasks already?}
    
    We present benchmarks for attacking detection and labeling.

    \item \textbf{Is there a repository that links to any or all papers or systems that use the dataset?} \eat{If so, please provide a link or other access point.}
    
    \citet{xie-etal-2021-models} use this dataset for inferring attributes of the attacking algorithms.

    \item \textbf{What (other) tasks could the dataset be used for?}
    
    TCAB can also be used for attack localization, attack target labeling, and attack characterization; see~\S\ref{sec:potential_tasks} for details.
    
    \item \textbf{Is there anything about the composition of the dataset or the way it was collected and preprocessed/cleaned/labeled that might impact future uses?} \eat{For example, is there anything that a future user might need to know to avoid uses that could result in unfair treatment of individuals or groups (e.g., stereotyping, quality of service issues) or other undesirable harms (e.g., financial harms, legal risks) If so, please provide a description. Is there anything a future user could do to mitigate these undesirable harms?}
    
    Currently, all instances are in English, and thus might yield observations that are specific to English. However, we publicly release our code for generating adversarial attacks and training victim models, enabling the extension of TCAB to other languages.

    \item \textbf{Are there tasks for which the dataset should not be used?} \eat{If so, please provide a description.}
    
    No.

    \eat{\item \textbf{Any other comments?} 
    
    No.}
    
\end{itemize}

\subsubsection*{Distribution}

\begin{itemize}

    \item \textbf{Will the dataset be distributed to third parties outside of the entity (e.g., company, institution, organization) on behalf of which the dataset was created?} \eat{If so, please provide a description.}
    
    The dataset is publicly available.

    \item \textbf{How will the dataset will be distributed (e.g., tarball on website, API, GitHub)?} Does the dataset have a digital object identifier (DOI)?
    
    All instances in TCAB are publicly available as CSV files on Zenodo~(\url{https://zenodo.org/record/7226519}). The DOI of the dataset is \textbf{10.5281/zenodo.7226519}.

    \item \textbf{When will the dataset be distributed?}
    
    The dataset is currently available at \url{https://react-nlp.github.io/tcab/}.

    \item \textbf{Will the dataset be distributed under a copyright or other intellectual property (IP) license, and/or under applicable terms of use (ToU)?} \eat{If so, please describe this license and/or ToU, and provide a link or other access point to, or otherwise reproduce, any relevant licensing terms or ToU, as well as any fees associated with these restrictions.}
    
    The dataset is distributed with a Creative Commons Attribution 4.0 International license.

    \item \textbf{Have any third parties imposed IP-based or other restrictions on the data associated with the instances?} \eat{If so, please describe these restrictions, and provide a link or other access point to, or otherwise reproduce, any relevant licensing terms, as well as any fees associated with these restrictions.}
    
    No.

    \item \textbf{Do any export controls or other regulatory restrictions apply to the dataset or to individual instances?} \eat{If so, please describe these restrictions, and provide a link or other access point to, or otherwise reproduce, any supporting documentation.}
    
    No.

    \eat{\item \textbf{Any other comments?}
    
    No.}

\end{itemize}

\subsubsection*{Maintenance}

\begin{itemize}

    \item \textbf{Who will be supporting/hosting/maintaining the dataset?} 
    
    The REACT-NLP team will continue to support the TCAB dataset.

    \item \textbf{How can the owner/curator/manager of the dataset be contacted (e.g., email address)?}
    
    Please direct any questions to Sameer Singh (sameer@uci.edu) or Daniel Lowd (lowd@cs.uoregon.edu).

    \item \textbf{Is there an erratum?} \eat{If so, please provide a link or other access point.}
    
    We will provide an erratum on the github repository if errors are found in the future.
    
    \item \textbf{Will the dataset be updated (e.g., to correct labeling errors, add new instances, delete instances)?} \eat{If so, please describe how often, by whom, and how updates will be communicated to users (e.g., mailing list, GitHub)?}
    
    If errors need to be corrected, we will publish a new version of the dataset and record those changes on the main project website:~\url{https://react-nlp.github.io/tcab/}.

    \item \textbf{If the dataset relates to people, are there applicable limits on the retention of the data associated with the instances (e.g., were individuals in question told that their data would be retained for a fixed period of time and then deleted)?} \eat{If so, please describe these limits and explain how they will be enforced.}
    
    Our dataset is not related to people.

    \item \textbf{Will older versions of the dataset continue to be supported/hosted/maintained?} \eat{If so, please describe how. If not, please describe how its obsolescence will be communicated to users.}
    
    If TCAB is updated in the future, we will ensure backward compatibility by time-stamping different versions of the dataset.

    \item \textbf{If others want to extend/augment/build on/contribute to the dataset, is there a mechanism for them to do so?} \eat{If so, please provide a description. Will these contributions be validated/verified? If so, please describe how. If not, why not? Is there a process for communicating/distributing these contributions to other users? If so, please provide a description.} 
    
    TCAB is designed to be extended with new datasets, attack instances, and target models. Code and instructions for extending TCAB is available at~\url{https://github.com/REACT-NLP/tcab_generation}.

    \eat{\item \textbf{Any other comments?}
    
    No.}

\end{itemize}

\subsection{Responsibility Statement}

The authors declare that they bear all responsibility for violations of rights related to this dataset.

\newpage
\section{TCAB Benchmark Details}
\label{app_sec:tcab_details}

Adversarial examples are generated using a TITAN RTX GPU with 24GB of memory and an Intel(R) Xeon(R) CPU E5-2690 v4 @ 2.6GHz with 60GB of memory. Attacks are run using Python 3.8. Dataset generation code is at~\url{https://github.com/REACT-NLP/tcab_generation}.

\subsection{Domain Datasets}
\label{app_sec:domain_datasets}

\begin{itemize}

    \item \textbf{Climate Change}\footnote{\url{https://www.kaggle.com/edqian/twitter-climate-change-sentiment-dataset}} consists of 62,356 tweets from Twitter pertaining to climate change. The collection of this data was funded by a Canada Foundation for Innovation JELF Grant to Chris Bauch, University of Waterloo. The goal is to determine if the text has a negative, neutral, or positive sentiment. On average, there are 17 words per instance. 

    \item \textbf{IMDB} consists of 50,000 highly polar movie reviews collected from \url{imdb.com} curated by~\citet{maas-EtAl:2011:ACL-HLT2011}. The goal is to determine if the text has a negative or positive sentiment. On average, there are 231 words per instance.

    \item \textbf{SST-2}~\cite{socher2013recursive} contains 68,221 phrases with fine-grained sentiment labels in the parse trees of 11,855 sentences from movie reviews. The goal is to determine if the text has a negative or positive sentiment. On average, there are 19 words per instance.

    \item \textbf{Wikipedia} (Talk Pages) consists of 159,686 comments~(9.6\% toxic) from Wikipedia editorial talk pages. The data was curated by~\citet{wulczyn2017ex} and made readily available by~\citet{dixon2018measuring}. The goal is to distinguish between non-toxic and toxic comments. On average, there are 68 words per instance.

    \item \textbf{Hatebase}~\cite{davidson2017automated} consists of 24,783~(83.2\% toxic) tweets from Twitter collected via searching for tweets containing words from the lexicon provided by \url{hatebase.org}. The goal is to distinguish between non-toxic and toxic comments. On average, there are 14 words per instance.

    \item \textbf{Civil Comments}\footnote{\url{https://www.kaggle.com/c/jigsaw-unintended-bias-in-toxicity-classification}} consists of 1,804,874~(8\% toxic) messages collected from the platform Civil Comments. The goal is to distinguish between non-toxic and toxic comments. On average, there are 51 words per instance.
\end{itemize}

\newpage
\subsection{Target Models}
\label{app_sec:target_models}

We use the popular HuggingFace transformers library\footnote{\url{https://huggingface.co/}} to fine-tune three transformer-based models designed for text classification~(BERT~\cite{devlin2019bert}, RoBERTa~\cite{liu2019roberta}, and XLNet~\cite{yang2019xlnet}) on each domain dataset. Table~\ref{tab:target_model_training_hyperparameters} shows the training parameters used to fine-tune/train each model. We use cross entropy as the loss function and Adam as the optimizer to train all models.

\begin{table*}[h]
\small
\setlength\tabcolsep{1.8pt}
\center
\begin{tabular}{llrrrrrr}
\toprule
\textbf{Dataset} & \textbf{Model} & \textbf{Max. len.} & \textbf{Learning rate} & \textbf{Batch size} & \textbf{No. epochs} & \textbf{Decay} & \textbf{Max. norm} \\
\midrule
\multirow{3}{*}{Climate Change}
    & BERT    & 250 & $1e^{-5}$ & 64 & 15 & 0 & 1.0 \\
    & RoBERTa & 250 & $1e^{-5}$ & 64 & 15 & 0 & 1.0 \\
    & XLNet   & 250 & $4e^{-5}$ & 64 & 15 & 0 & 1.0 \\
\midrule
\multirow{3}{*}{IMDB}
    & BERT    & 128 & $4e^{-5}$ & 64 & 5  & 0 & 1.0 \\
    & RoBERTa & 128 & $1e^{-6}$ & 64 & 10 & 0 & 1.0 \\
    & XLNet   & 128 & $4e^{-5}$ & 64 & 5  & 0 & 1.0 \\
\midrule
\multirow{3}{*}{SST-2}
    & BERT    & 128 & $1e^{-5}$ & 32 & 5 & 0 & 1.0 \\
    & RoBERTa & 128 & $1e^{-5}$ & 32 & 5 & 0 & 1.0 \\
    & XLNet   & 128 & $1e^{-5}$ & 32 & 5 & 0 & 1.0 \\
\midrule
\multirow{3}{*}{Wikipedia}
    & BERT    & 250 & $1e^{-6}$ & 32 & 10 & 0 & 1.0 \\
    & RoBERTa & 250 & $1e^{-6}$ & 32 & 10 & 0 & 1.0 \\
    & XLNet   & 250 & $1e^{-6}$ & 16 & 10 & 0 & 1.0 \\
\midrule
\multirow{3}{*}{Hatebase}
    & BERT    & 250 & $1e^{-6}$ & 32 & 50 & 0 & 1.0 \\
    & RoBERTa & 250 & $1e^{-6}$ & 32 & 50 & 0 & 1.0 \\
    & XLNet   & 128 & $1e^{-6}$ & 16 & 50 & 0 & 1.0 \\
\midrule
\multirow{3}{*}{Civil Comments}
    & BERT    & 250 & $1e^{-6}$ & 32 & 10 & 0 & 1.0 \\
    & RoBERTa & 250 & $1e^{-6}$ & 32 & 10 & 0 & 1.0 \\
    & XLNet   & 128 & $1e^{-6}$ & 16 & 10 & 0 & 1.0 \\
\bottomrule
\end{tabular}
\caption{Training parameters used to fine-tune/train each target model. Max. len. is the maximum number of tokens fed into each model; Decay denotes the weight decay of the model.}
\label{tab:target_model_training_hyperparameters}
\end{table*}

\newpage
\subsection{Attack Methods}
\label{app_sec:attack_methods}

In Table \ref{tab:attack_methods}, all of the attack methods used to create TCAB are listed along with the toolchain each attack method belongs to and the access level, linguistic constraints, and perturbation level each attack method has. Linguistic constraints ensure coherency of perturbed sentences. Such constraints include semantic similarity between the original and perturbed text evaluated via a language model on the word or sentence level, distance metrics in word embedding space, or restricting perturbations to sememes, for example.

\begin{table}[ht]
\center
\caption{Attack methods used to create TCAB.}
\begin{tabular}{lccm{3.5cm}c}
\toprule
\textbf{Attack method} & \textbf{Toolchain} & \textbf{Access level} & \textbf{Linguistic constraints} & \textbf{Perturb. level} \\
\midrule
BAE~\cite{ramakrishnan2020bae} & TextAttack & Black box & USE~\cite{cer2018universal} cosine similarity & Word \\
\addlinespace
DeepWordBug~\cite{gao2018deepwordbug} & TextAttack & Gray box  & Levenshtein edit distance & Char \\
\addlinespace
FasterGenetic~\cite{jia2019certified} & TextAttack & Gray box  & Percentage of words perturbed, Language Model perplexity, Word embedding distance & Word \\
\addlinespace
Genetic~\cite{alzantot2018genetic} & OpenAttack & Gray box  & Percentage of words perturbed, Language Model perplexity, Word embedding distance & Word \\
\addlinespace
HotFlip~\cite{ebrahimi2018hotflip} & OpenAttack & White box & Word Embedding Cosine Similarity, Part-of-speech match, Number of words perturbed & Char \\
\addlinespace
IGA~\cite{wang2019igawang} & TextAttack & Gray box  & Percentage of words perturbed, Word embedding distance & Word \\
\addlinespace
Pruthi~\cite{pruthi2019combating} & TextAttack & Gray box & Minimum word length, Maximum number of words perturbed & Word \\
\addlinespace
PSO~\cite{zang2020pso} & TextAttack & Black box & Sememes only & Word \\
\addlinespace
TextBugger~\cite{li2018textbugger} & TextAttack & Black box & USE~\cite{cer2018universal} cosine similarity & Char \\
\addlinespace
TextBugger~\cite{li2018textbugger} & OpenAttack & White box & USE~\cite{cer2018universal} cosine similarity & Word+Char \\
\addlinespace
TextFooler~\cite{jin2019textfooler} & TextAttack & Black box & Word Embedding Distance, Part-of-speech match, USE sentence encoding cosine similarity & Word \\
\addlinespace
VIPER~\cite{eger2019viper} & OpenAttack & Black box & Visual embedding space & Char \\
\bottomrule
\end{tabular}
\label{tab:attack_methods}
\end{table}

\newpage
\subsection{Attack Samples}
\label{app_sec:attack_samples}

Tables \ref{tab:attack_samples_sst}, \ref{tab:attack_samples_wikipedia_1} and \ref{tab:attack_samples_wikipedia_2} show successful adversarial attacks from multiple methods on the SST and Wikipedia datasets with XLNet as the target model. Table \ref{tab:attack_samples_sst_multi} shows successful BAE attacks on SST with RoBERTa as the target model.

\begin{table}[ht]
\center
\caption{Successful attacks on SST with XLNet as the target model.}
\label{tab:attack_samples_sst}
\begin{tabular}{lm{7.25cm}cr}
\toprule
\textbf{Attack} & \textbf{Text} & \textbf{Label} & \textbf{Confidence}\\
\midrule
    Original
    & Part comedy, part drama, the movie winds up accomplishing neither in full, and leaves us feeling touched and amused by several moments and ideas, but nevertheless dissatisfied with the movie as a whole.
    & Negative 
    & 89.2\% \\
\midrule
    BAE
    & Part comedy, part drama, the movie winds up accomplishing neither in full, and leaves us feeling touched and amused by several moments and ideas, but nevertheless \textcolor{red}{satisfied} with the movie as a whole.
    & Positive 
    & 84.4\%\\
\midrule
    DeepWordBug
    &Part comedy, part drama, the movie wi\textcolor{red}{r}nds up accomplishing \textcolor{red}{en}ither in full, and leaves us feeling touched and amused by several moments and ideas, but nevertheless dissat\textcolor{red}{A}sfied with the movie as a whole.
    & Positive 
    & 60.1\%\\
\midrule
    FasterGenetic
    & Part comedy, part drama, the movie winds up accomplishing \textcolor{red}{or} in full, and leaves us feeling touched and amused by several moments and ideas, but \textcolor{red}{notwithstanding displeased} with the movie as a whole.  
    & Positive 
    & 96.2\%\\
\midrule
    PSO
    & Part comedy, part drama, the movie winds up accomplishing neither in full, and leaves us feeling touched and amused by several moments and ideas, but nevertheless \textcolor{red}{dazzling} with the movie as a whole.
    & Positive 
    & 94.8\%\\
\midrule
    Pruthi
    & Part comedy, part drama, the movie winds up accomplishing neither in full, and leaves us feeling touched and amused by several moments and ideas, but neverthel\textcolor{red}{w}ss dissatisfied with the movie as a whole.  
    & Positive 
    & 95.5\%\\
\midrule
    TextFooler
    & Part comedy, part drama, the movie \textcolor{red}{gusty} up accomplishing neither in full, and leaves us feeling touched and amused by several moments and ideas, but nevertheless \textcolor{red}{incensed} with the movie as a whole. 
    & Positive 
    & 74.1\%\\
\midrule
    Genetic
    &  \textcolor{red}{Portions} comedy, part \textcolor{red}{theatrical}, the movie winds up \textcolor{red}{achieving nor} in full, and leaves us feeling touched and \textcolor{red}{tickled} by \textcolor{red}{diverse} moments and \textcolor{red}{thought}, but \textcolor{red}{however} dissatisfied with the movie as a whole.
    & Positive 
    & 58.3\%\\
\midrule
    TextBugger
    & Part comedy, part drama, the movie winds up accomplishing \textcolor{red}{nor} in full, and leaves us feeling touched and amused by several moments and ideas, but nevertheless \textcolor{red}{‘Å}issatisfied with the movie as a whole. 
    & Positive 
    & 53.3\%\\
\bottomrule
\end{tabular}
\end{table}

\newpage

\begin{table}[ht]
\caption{Successful adversarial attacks on Wikipedia with XLNet as the target model.}
\label{tab:attack_samples_wikipedia_1}
\center
\begin{tabular}{lm{6.5cm}cr}
\toprule
\textbf{Attack} & \textbf{Text} & \textbf{Label} & \textbf{Confidence}\\
\midrule
    Original
    & YOu moron, do not accuse me to go and study. The fools and used like condom people with small or with no brains are YOU! \dots
    & Toxic
    & 98.3 \%\\
\midrule
    BAE
    & YOu \textcolor{red}{suke}, do not accuse me to go and study. The fools and used like condom people with small or with no brains are YOU! \dots
    & Non-toxic
    & 78.5\%\\
\midrule
    DeepWordBug
    & \textcolor{red}{Yu mBron}, do not accuse me to go and study. The fools and used like condom people with small or with no brains are YOU! \dots
    & Non-toxic
    & 76.0\%\\
\midrule
    TextBugger(TA)
    & YOu \textcolor{red}{moon}, do not accuse me to go and study. The fools and used like condom people with small or with no brains are YOU! \dots
    & Non-toxic
    & 68.4\%\\
\midrule
    TextFooler
    & \textcolor{red}{tu cretin}, do not accuse me to go and study. The fools and used like condom people with small or with no brains are YOU! \dots
    & Non-toxic
    & 67.2\%\\
\midrule
    TextBugger(OA)
    &  Y\textcolor{red}{o}u moron , do not accuse me to go and study\textcolor{red}{ing}. The f\textcolor{red}{o}ls and \textcolor{red}{utilized loves} condom people with little or with no brains \textcolor{red}{z}re YOU!  \dots
    & Non-toxic
    & 91.1\%\\
\midrule
    Genetic
    & Y\textcolor{red}{o}u \textcolor{red}{cretin} , do not accuse me to go and study. The fools and \textcolor{red}{usage} like condom people with small or with no brains are \textcolor{red}{you}! \dots
    & Non-toxic
    & 54.4\%\\
\bottomrule
\end{tabular}
\end{table}

\newpage

\begin{table}[h!]
\caption{More successful adversarial attacks on Wikipedia with XLNet as the target model.}
\label{tab:attack_samples_wikipedia_2}
\center
\begin{tabular}{lm{6.5cm}cr}
\toprule
\textbf{Attack} & \textbf{Text} & \textbf{Label} & \textbf{Confidence}\\
\midrule
    Original
    & Why do not the two of you, the mentally r***ded Jennygirl and the omnipotent Richie Rich, go together to Harrods and \dots when you come back after they would be kicking off your dipshit spinning incapable ass from the central perk. What a pity. Receive all my sympathy at one time. \dots
    & Toxic
    & 63.7 \%\\
\midrule
    Pruthi
    & Why do not the two of you, the mentally \textcolor{red}{regarded} Jennygirl and the omnipotent Richie Rich, go together to Harrods and \dots when you come back after they would be kicking off your dipshit spinning incapable ass from the central perk. What a pity. Receive all my sympathy at one time.\dots
    & Non-toxic
    & 82.1\%\\
\midrule
    PSO
    & Why do not the two of you, the mentally \textcolor{red}{checked} Jennygirl and the omnipotent Richie Rich, go together to Harrods and \dots when you come back after they would be kicking off your dipshit spinning incapable ass from the central perk. What a pity. Receive all my sympathy at one time. \dots
    & Non-toxic
    & 85.7\%\\
\midrule
    BAE
    & Why do not the two of you, the mentally r***ded Jennygirl and the omnipotent Richie Rich, go together to Harrods and \dots when you come back after they would be kicking off your \textcolor{red}{rella} spinning incapable ass from the central perk. What a pity. Receive all my sympathy at one time. \dots
    & Non-toxic
    & 92.2\%\\
\midrule
    TextFooler
    & Why do not the two of you, the mentally r***ded Jennygirl and the omnipotent Richie Rich, go together to Harrods and \dots when you come back after they would be kicking off your \textcolor{red}{cretin} spinning incapable ass from the central perk. What a pity. Receive all my sympathy at one time. \dots
    & Non-toxic
    & 91.2\%\\
\midrule
    Genetic
    &  Why do not the two of you, the mentally \textcolor{red}{backward} Jennygirl and the omnipotent Richie Rich, go together to Harrods and \dots when you come back after they would be kicking off your dipshit spinning incapable ass from the central perk. What a pity. Receive all my sympathy at one time. \dots
    & Non-toxic
    & 62.3\%\\
\midrule
    DeepWordBug
    &  Why do not the two of you, the mentally r***ded Jennygirl and the omnipotent Richie Rich, go together to Harrods and \dots when you come back after they would be kicking off your dips\textcolor{red}{g}it spinning incapable ass from the central perk. What a pity. Receive all my sympathy at one time. \dots
    & Non-toxic
    & 93.7\%\\
\bottomrule
\end{tabular}
\end{table}

\newpage

\begin{table}[h!]
\caption{Successful BAE attacks on SST with RoBERTa as the target model. The ``Original" column contains the original label and prediction confidence. The ``Perturbed" column contains the label prediction and confidence for the perturbed text.}
\label{tab:attack_samples_sst_multi}
\center
\begin{tabular}{m{4.6cm}m{1.3cm}m{4.6cm}m{1.4cm}}
\toprule
\textbf{Original Text} & \textbf{Original} & \textbf{Perturbed Text} & \textbf{Perturbed}\\
\midrule
    Watching Trouble Every Day , at least if you do n't know what 's coming , is like biting into what looks like a juicy , delicious plum on a hot summer day and coming away with your mouth full of rotten pulp and living worms .
    & Negative (92.6\%)
    & Watching Trouble Every Day , at least if you do n't know what 's coming , is like biting into what looks like a juicy , delicious plum on a hot summer day and coming away with your mouth full of \textcolor{red}{fresh} pulp and living worms .
    & Positive (55.3\%)\\
\midrule
    With a story inspired by the tumultuous surroundings of Los Angeles , where feelings of marginalization loom for every dreamer with a burst bubble , The Dogwalker has a few characters and ideas , but it never manages to put them on the same path .
    & Negative (91.1\%)
    & With a story inspired by the tumultuous surroundings of Los Angeles , where feelings of marginalization loom for every dreamer with a burst bubble , The Dogwalker has a few characters and ideas , but it \textcolor{red}{still} manages to put them on the same path .
    & Positive (98.2\%) \\
\midrule
    To imagine the life of Harry Potter as a martial arts adventure told by a lobotomized Woody Allen is to have some idea of the fate that lies in store for moviegoers lured to the mediocrity that is Kung Pow : Enter the Fist .
    & Negative (94.0\%)
    & To \textcolor{red}{experience} the life of Harry Potter as a martial arts adventure told by a \textcolor{red}{mad} Woody Allen is to have some idea of the fate that lies in store for moviegoers lured to the mediocrity that is Kung Pow : Enter the Fist .
    & Positive (51.1\%)\\
\midrule
    This is n't a narrative film -- I do n't know if it 's possible to make a narrative film about September 11th , though I 'm sure some will try -- but it 's as close as anyone has dared to come .
    & Positive, (63.8\%)
    & This is n't a narrative film -- I \textcolor{red}{do t} know if it 's possible to make a narrative film about September 11th , though I 'm sure some will try -- but it 's as close as anyone has dare to come .
    & Negative (66.6\%)\\
\midrule
    Though Mama takes a bit too long to find its rhythm and a third-act plot development is somewhat melodramatic , its ribald humor and touching nostalgia are sure to please anyone in search of a Jules and Jim for the new millennium .
    & Positive (98.2\%)
    & \textcolor{red}{as} Mama takes a bit too long to find its rhythm and a third-act plot development is somewhat \textcolor{red}{questionable} , its ribald humor and touching nostalgia are sure to \textcolor{red}{deter} anyone in search of a Jules and Jim for the new millennium .
    & Negative (54.7\%)\\
\bottomrule
\end{tabular}
\end{table}

\newpage
\subsection{Human Evaluation Details}
\label{app_sec:human_eval}

Here we provide additional details about the human evaluation on a portion of TCAB.

\subsubsection*{User Interfaces}
\label{app_sec:UI}
\begin{figure*}[h]
\centering
    \begin{minipage}{\textwidth}
        \centering
        \begin{framed}
        \includegraphics[width=\textwidth]{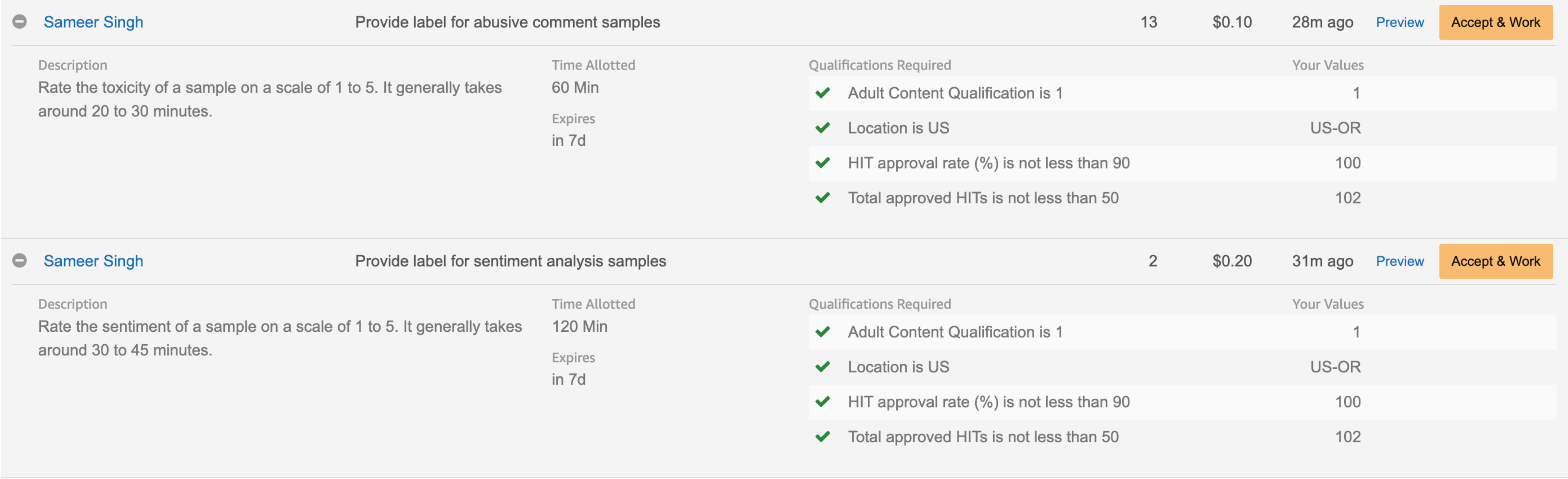}
        \end{framed}
        \caption{The requirements for MTurk workers and the appearance of HITs}
        \label{fig:human_eval_requirements}
    \end{minipage}
\end{figure*}

Figure~\ref{fig:human_eval_requirements} presents the requirements for MTurk workers with some additional details, such as the price per HIT, and the estimated time for the task. 

\begin{figure}[h]
\centering
    \begin{minipage}{\textwidth}
        \centering
        \begin{framed}
        \includegraphics[width=\textwidth]{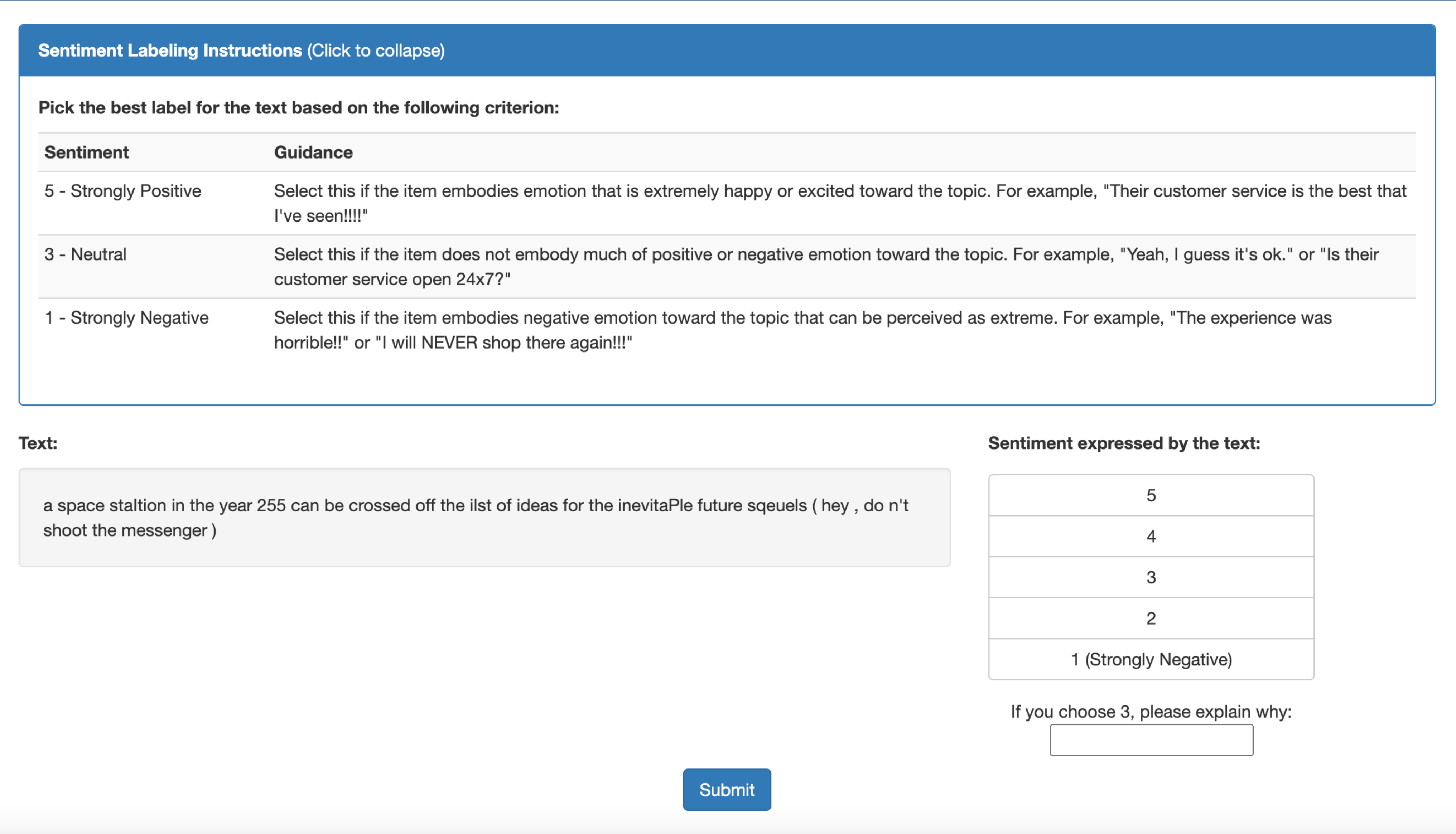}
        \end{framed}
        \caption{The evaluation interface for sentiment analysis}
        \label{fig:sentiment_UI}
    \end{minipage}
\end{figure}

\begin{figure}[h]
\centering
    \begin{minipage}{\textwidth}
        \centering
        \begin{framed}
        \includegraphics[width=\textwidth]{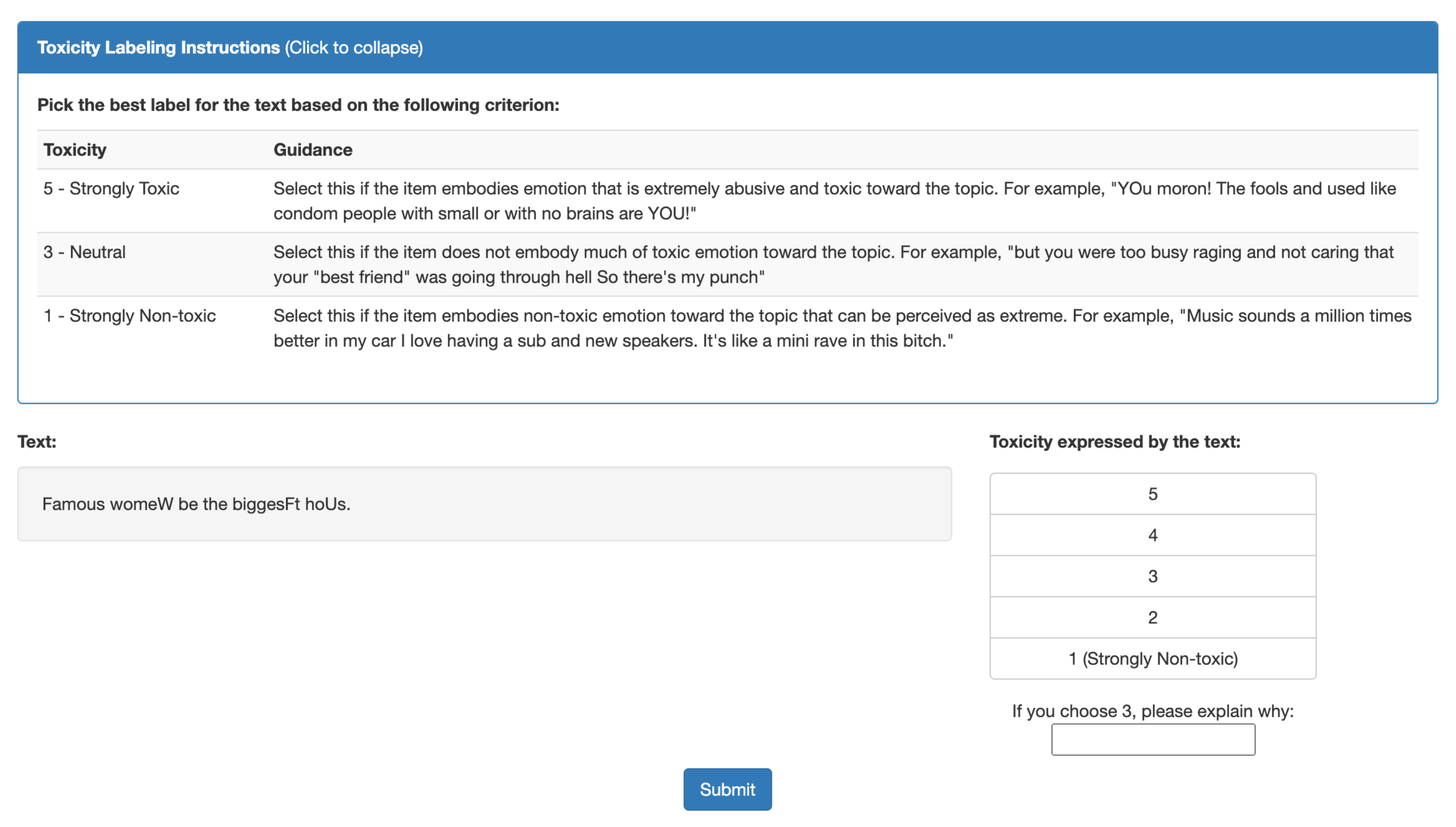}
        \end{framed}
        \caption{The evaluation interface for abuse detection}
        \label{fig:abuse_UI}
    \end{minipage}
\end{figure}

We wrote the user interfaces for MTurk using HTML. Screenshots are shown in Figures~\ref{fig:sentiment_UI}, and~\ref{fig:abuse_UI}. A brief instruction of the task with a few examples are provided at the top of the page. The text to be evaluated is presented at the bottom left. The scores are presented on the right side, followed by a text input box for additional comments.

\newpage

\subsubsection*{Compensation}
\label{app_sec:compensation}

In TCAB, we implemented $3$ target models, $6$ datasets, and $12$ attacks. For every target model/dataset/attack combination, approximately, we compensate workers:
\begin{itemize}
    \item IMDB: $0.2$ USD per HIT. Every HIT is evaluated by $5$ unique workers. Each worker evaluates about $15$ unique HITs.
    \item All other datasets: $0.1$ USD per HIT. Every HIT is evaluated by $5$ unique workers. Each worker evaluates about $30$ unique HITs.
\end{itemize}
This compensation rate was chosen to ensure an overall pay rate of at least \$15/hour for all workers. Therefore, we collected roughly $29,700$ unique evaluation results from MTurk for all target model/dataset/attack combinations. The total cost was roughly \$3,200 USD.

\paragraph{Detailed Results.}

Table~\ref{tab:human_eval_attacks} shows the percentage of labels preserved broken down by each attack.

\begin{table}[ht]
\center
\caption{Percentage of adversarial instances whose labels were preserved after human evaluation for each attack method on each dataset. "-" means no successful attacks for that method on that dataset.}
\label{tab:human_eval_attacks}
\begin{tabular}{lcccccc}
\toprule
\textbf{Attack} & \textbf{Climate Change} & \textbf{SST-2} & \textbf{IMDB} & \textbf{Hatebase} & \textbf{Wikipedia} & \textbf{Civil Comments} \\
\midrule
    \textbf{BAE} & 48 & 31 & 45 & 78 & 79 & 74 \\
    \textbf{DWB} & 48 & 58 & 80 & 87 & 84 & 84 \\
    \textbf{FG}  & 44 & 48 & 57 & 74 & 79 & 82 \\
    \textbf{Gn.*} & 58 & 47 & 53 & 84 & 81 & 84 \\
    \textbf{HF*} & 54 & 51 & 45 & 77 & 83 & 81 \\
    \textbf{IGA} & 53 & 62 & - & 84 & - & 83 \\
    \textbf{Pr.} & 57 & 61 & 50 & 84 & 78 & 79 \\
    \textbf{PSO} & 52 & 52 & 49 & 81 & 73 & 78 \\
    \textbf{TB*} & 57 & 50 & 46 & 77 & 88 & 83 \\
    \textbf{TB} & 54 & 67 & 56 & 79 & 89 & 85 \\
    \textbf{TF} & 54 & 43 & 41 & 82 & 74 & 80 \\
    \textbf{VIP*} & 53 & 54 & 35 & 82 & 87 & 80 \\
\bottomrule
\end{tabular}
\end{table}

\newpage

\section{Baseline Model Details}
\label{app_sec:baseline_models}

Here we provide details on the baseline models we use for attack detection and labeling. Baseline model code is at~\url{https://github.com/REACT-NLP/tcab_benchmark}.

\subsection{Hand-Crafted Features}
\label{app_sec:engineered_features}

This section provides a detailed description of each feature for the text, language model, and target model properties.

\subsubsection*{Text Properties.}

\begin{itemize}[nosep]
    \item \textbf{BERT features}: BERT embedding representation of the input sequence.

    \item \textbf{No. chars.}: Number of characters.

    \item \textbf{No. alpha chars.}: Number of alphabet characters (a-z).

    \item \textbf{No. digit chars.}: Number of digit characters (0-9).

    \item \textbf{No. punctuation.}: Number of punctuation mark characters (“?”, “!”, etc.)

    \item \textbf{No. multi. spaces.}: Number of times multiple spaces appear between words.

    \item \textbf{No. words.}: Number of words.

    \item \textbf{Avg. word len.}: Mean, variance, and quantiles (25\%, 50\%, and 75\%) of the number of characters per word for different regions of the input (first 25\%, middle 50\%, last 25\%, entire input).

    \item \textbf{No. non-ascii.}: Number of non-ascii characters.

    \item \textbf{Cased letters.}: Number of uppercase letters, number of lowercase letters, fraction of uppercase letters, and fraction of lowercase letters.

    \item \textbf{Is first word lowercase.}: True if the first character of the first word is lowercase, otherwise False.

    \item \textbf{No. mixed-case words.}: Number of words that contain both upper and lowercase letters (not including the first letter of each word).

    \item \textbf{No. single lowercase letters.}: Number of single-letter-lowercase words (excluding “a” and “i”).

    \item \textbf{No. lowercase after punctuation.}: Number of words after a punctuation that begin with a lowercase letter.

    \item \textbf{No. cased word switches.}: Number of times words switch from all uppercase to all lowercase and vice versa (e.g. THIS IS a sentence THAT contains 3 switches).
\end{itemize}

\subsubsection*{Language Model Properties.}

\begin{itemize}[nosep]
    \item \textbf{Probability and rank}: Mean, variance, and quantiles (25\%, 50\% and 75\%) of the token probabilities and ranks for different regions of the input (first 25\%, middle 50\%, last 25\%, entire input) using RoBERTa, a masked language model.

    \item \textbf{Perplexity}: Perplexity for different regions of the input (first 25\%, middle 50\%, last 25\%, entire input) using GPT-2, a causal language model.
\end{itemize}

\subsubsection*{Target Model Properties.}

For the following properties, we assume the target model to be a RoBERTa text classifier.

\begin{itemize}[nosep]
    \item \textbf{Posterior.}: Output posteriors of the target model (softmax applied to logits).

    \item \textbf{Gradient.} Mean, variance, and quantiles (25\%, 50\%, and 75\%) of the gradients for each layer of the target model given different regions of the input (first 25\%, middle 50\%, last 25\%, entire input).

    \item \textbf{Activation.}: Mean, variance, and quantiles (25\%, 50\%, and 75\%) of the node activations for each layer of the target model given different regions of the input (first 25\%, middle 50\%, last 25\%, entire input).

    \item \textbf{Saliency.}: Mean, variance, and quantiles (25\%, 50\%, and 75\%) of the saliency values (gradients of the target model with respect to the input tokens) given the input.
\end{itemize}

\newpage
\subsection{Baseline Model Overview}

\begin{figure}[h]
    \centering
    \includegraphics[width=0.75\textwidth]{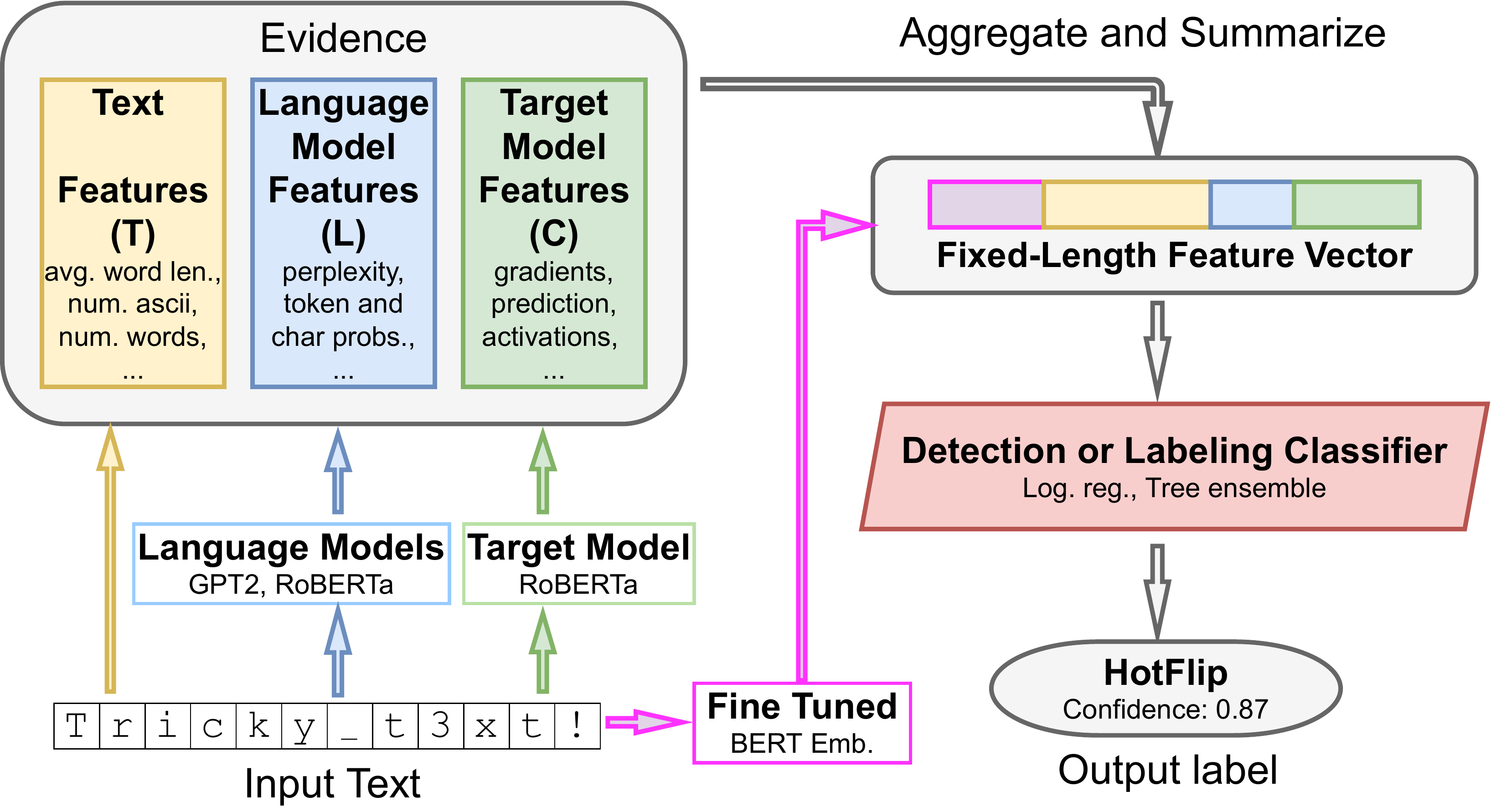}
    \caption{Baseline model for attack detection and labeling.}
    \label{fig:experiment_pipeline}
\end{figure}

Figure~\ref{fig:experiment_pipeline} shows an overview of our baseline approach for detecting and identifying attacks in TCAB.

\subsection{Additional Baseline Results}
\label{app_sec:baseline_results}

Tables~\ref{tab:baselines_bert}~and~\ref{tab:baselines_xlnet} show attack detection and labeling results for attacks targeting BERT and XLNet, respectively.

\begin{table}[h]
\small
\setlength\tabcolsep{3pt}
\center
\caption{Attack detection and labeling results showing the balanced accuracy of each baseline model on each dataset for attacks targeting BERT.}
\label{tab:baselines_bert}
\begin{tabular}{l ccc c ccc}
\toprule
\multirow{2}{*}{}
  & \multicolumn{3}{c}{\textbf{Attack Detection} (2 classes)}
  &
  & \multicolumn{3}{c}{\textbf{Attack Labeling} (12 classes)} \\
\cmidrule(lr){2-4}\cmidrule(lr){6-8}
 \textbf{Dataset} &
 FT &
 LR\scriptsize-FT-TLC &
 LGB\scriptsize-FT-TLC &
 & FT &
 LR\scriptsize-FT-TLC &
 LGB\scriptsize-FT-TLC \\
\midrule
Climate Change
 & 87.0 & 88.2 & \textbf{88.4}
 &
 & 55.0 & 66.1 & \textbf{67.2}
\\
IMDB
 & 91.8 & 92.6 & \textbf{93.8}
 &
 & 70.3 & 74.3 & \textbf{79.0}
\\
SST-2
 & \textbf{93.7} & 91.1 & 92.8
 &
 & 57.4 & \textbf{63.4} & 63.0
\vspace{0.01cm} \\
\hdashline \\
\vspace{-0.6cm} \\
Wikipedia
 & 87.6 & 89.0 & \textbf{91.4}
 &
 & 42.2 & 46.4 & \textbf{49.1}
\\
Hatebase
 & 91.7 & 92.6 & \textbf{93.6}
 &
 & 61.1 & 63.6 & \textbf{63.7}
\\
Civil Comments
 & 86.4 & 88.4 & \textbf{89.2}
 &
 & 58.9 & 60.9 & \textbf{62.6}
\\
    \bottomrule
    \end{tabular}
\end{table}

\begin{table}[h]
\small
\setlength\tabcolsep{3pt}
\center
\caption{Attack detection and labeling results showing the balanced accuracy of each baseline model on each dataset for attacks targeting XLNet.}
\label{tab:baselines_xlnet}
\begin{tabular}{l ccc c ccc}
\toprule
\multirow{2}{*}{}
  & \multicolumn{3}{c}{\textbf{Attack Detection} (2 classes)}
  &
  & \multicolumn{3}{c}{\textbf{Attack Labeling} (12 classes)} \\
\cmidrule(lr){2-4}\cmidrule(lr){6-8}
 \textbf{Dataset} &
 FT &
 LR\scriptsize-FT-TLC &
 LGB\scriptsize-FT-TLC &
 & FT &
 LR\scriptsize-FT-TLC &
 LGB\scriptsize-FT-TLC \\
\midrule
Climate Change
 & 89.0 & 89.0 & \textbf{89.8}
 &
 & 78.4 & 79.7 & \textbf{80.3}
\\
IMDB
 & 90.3 & 91.5 & \textbf{92.7}
 &
 & 64.3 & 72.8 & \textbf{78.2}
\\
SST-2
 & \textbf{92.5} & 91.9 & 92.3
 &
 & 59.2 & 63.9 & \textbf{66.1}
\vspace{0.01cm} \\
\hdashline \\
\vspace{-0.6cm} \\
Wikipedia
 & 87.8 & 88.7 & \textbf{90.7}
 &
 & 48.8 & 50.5 & \textbf{55.3}
\\
Hatebase
 & 88.6 & 88.9 & \textbf{89.3}
 &
 & 48.9 & 60.5 & \textbf{62.1}
\\
Civil Comments
 & 87.7 & 88.2 & \textbf{89.3}
 &
 & 60.3 & 61.9 & \textbf{63.7}
\\
    \bottomrule
    \end{tabular}
\end{table}

\newpage
\section{Additional Details}
\label{app_sec:additional_details}

Here we discuss any limitations or potential negative societal impacts of our work.

\subsection{Limitations}
\label{app_sec:limitations}

Different attack methods differ significantly in their attack success rates, making it more difficult to obtain enough adversarial examples for some attack methods than others. Also, we attack instances from the test set of each domain dataset, but one may obtain even more adversarial examples by training the target model on less data and attacking the unused parts of the training set; however, the target model should still have enough training data to train an accurate model worth attacking.

Many of the attacks in our dataset change the ``true'' label of the instances while changing the predicted label. This allows for greater scalability, but it also means that a substantial fraction of the attack instances are arguably not entirely successful. A filtered version of TCAB, which contained only human-verified attack instances, would be beneficial for understanding successful adversarial attacks but much more expensive to create.

\subsection{Potential Negative Societal Impacts}
\label{app_sec:impact}

Our work generally promotes learning as much as possible about existing and new attackers, and subsequently informing new design choices for defenses against attacks targeting text classifiers. Attackers might be able to use our dataset to develop better attacks that are harder to detect or characterize. Furthermore, some attacks could be socially beneficial, such as obfuscating text for privacy or to avoid political oppression. In these contexts, improved text classifiers and improved defenses against text attacks could cause some harm.

\end{document}